\documentclass[twoside,11pt]{article}

\usepackage{etoolbox}

\newbool{isarxiv}
\setbool{isarxiv}{true} 

\ifbool{isarxiv}{\usepackage{jmlr2eModifiedForArXiv}}{\usepackage{jmlr2e}}

\usepackage{algorithm,algorithmic}
\usepackage{wrapfig} 
\usepackage{sidecap}  
\usepackage{pgf,tikz,pgfplots}
\usetikzlibrary{arrows,automata,fit,positioning}
\usepackage{amssymb} 
\usepackage{amsfonts}
\usepackage{bm,amsbsy} 
\usepackage{amsmath}
\usepackage{stmaryrd}
\usepackage[normalem]{ulem}
\useunder{\uline}{\ul}{}
\usepackage{booktabs}
\usepackage{multirow}
\usepackage{tabularx}
\usepackage{caption}

\newcommand{\figref}[1]{Fig.~\ref{fig:#1}}  
\newcommand{\secref}[1]{Sec.~\ref{#1}}  
\newcommand{\algoref}[1]{Algorithm~\ref{algo:#1}}  
\newcommand{\defref}[1]{Def.~\ref{def:#1}}  

\newcommand{\vx}{\mathbf{x}}

\newcommand{\Dat}{\mathcal{D}}

\newcommand{\Nrm}{\mathcal{N}}
\newcommand{\trp}{{^\top}} 

\newcommand{\vphi}{\mathbf{\ensuremath{\bm{\phi}}}}

\newcommand{\vone}{\mathbf{1}} 
\newcommand{\vz}{\mathbf{z}}
\newcommand{\vw}{\mathbf{w}}
\newcommand{\vn}{\mathbf{n}}
 
\newcommand{\vl}{\mathbf{\ensuremath{\bm{\mathit{l}}}}}
\newcommand{\vm}{\mathbf{\ensuremath{\bm{\mathit{m}}}}} 
\newcommand{\vs}{\mathbf{s}}
\newcommand{\vb}{\mathbf{b}} 
\newcommand{\vc}{\mathbf{c}}

\newcommand{\vbeta}{\mathbf{\ensuremath{\bm{\beta}}}}
\newcommand{\vmu}{\mathbf{\ensuremath{\bm{\mu}}}}
\newcommand{\vgamma}{\mathbf{\ensuremath{\bm{\gamma}}}}
\newcommand{\vlambda}{\mathbf{\ensuremath{\bm{\lambda}}}}
\newcommand{\vtheta}{\mathbf{\ensuremath{\bm{\theta}}}}
\newcommand{\vnu}{\mathbf{\ensuremath{\bm{\nu}}}}
\newcommand{\vxi}{\mathbf{\ensuremath{\bm{\xi}}}}
\newcommand{\vpsi}{\mathbf{\ensuremath{\bm{\psi}}}}
\newcommand{\vzeta}{\mathbf{\ensuremath{\bm{\zeta}}}}

\newcommand{\vy}{\mathbf{y}}

\def\bbE{\mathbb{E}}

\newcommand{\mc}[1]{\mathcal{#1}}
\newcommand{\mb}[1]{\mathbf{#1}}

\newcommand{\appropto}{\mathrel{\vcenter{
  \offinterlineskip\halign{\hfil$##$\cr
    \propto\cr\noalign{\kern2pt}\sim\cr\noalign{\kern-2pt}}}}}

\renewcommand{\eqref}[1]{Eq.~\ref{eq:#1}}

\usepackage{todonotes}

\jmlrheading{1}{2016}{1-48}{4/00}{10/00}{Everybody}

\ShortHeadings{Variational Bayes In Private Settings (VIPS)}{Park Foulds Chaudhuri and Welling}
\firstpageno{1}

\begin{document}

\title{Variational Bayes In Private Settings (VIPS)}
\vspace{-0.2cm}
\author{\name Mijung Park \email mijungi.p@gmail.com \\
       \addr Max Planck Institute for Intelligent Systems \& University of T\"ubingen \\
       Max-Planck-Ring 4, 72076 T\"ubingen, Germany\\
       \name James Foulds  
       \email jfoulds@umbc.edu\\
       \addr Department of Information Systems, University of Maryland, Baltimore County.\\
       ITE 447, 1000 Hilltop Circle, Baltimore, MD 21250, USA\\
       \name Kamalika Chaudhuri \email kamalika@cs.ucsd.edu\\
       \addr Department of Computer Science, University of California, San Diego.\\
       EBU3B 4110,  University of California, San Diego, CA 92093, USA\\
       \name Max Welling \email m.welling@uva.nl\\
       \addr QUVA lab, Informatics Institute, University of Amsterdam\\
        Science Park 904, Amsterdam 1098 XH, The Netherlands
       }

\editor{}

\maketitle
\vspace{-0.2cm}

\begin{abstract}
Many applications of Bayesian data analysis involve sensitive information, motivating methods which ensure that privacy is protected.
We introduce a general privacy-preserving framework for Variational Bayes (VB), a widely used optimization-based Bayesian inference method.
Our framework respects differential privacy, the gold-standard privacy criterion, and encompasses a large class of probabilistic models, called the {\it{Conjugate Exponential}} (CE) family. 
We observe that we can straightforwardly privatise VB's approximate posterior distributions for models in the CE family, by perturbing the expected sufficient statistics of the complete-data likelihood.
For a broadly-used class of non-CE models, those with binomial likelihoods, we show how to bring such models into the CE family, such that inferences in the modified model resemble the private variational Bayes algorithm as closely as possible, using the P{\'o}lya-Gamma data augmentation scheme.  
The iterative nature of variational Bayes presents a further challenge since iterations increase the amount of noise needed.
We overcome this by combining: (1) an improved composition method for differential privacy, called the {\it{moments accountant}}, which provides a tight bound on the privacy cost of multiple VB iterations and thus significantly decreases the amount of additive noise;
and (2) the privacy amplification effect of subsampling mini-batches from large-scale data in stochastic learning.   
We empirically demonstrate the effectiveness of our method in CE and non-CE models
including latent Dirichlet allocation, Bayesian logistic regression,
and sigmoid belief networks, 
evaluated on real-world datasets. 
\end{abstract}

\begin{keywords}
Variational Bayes, differential privacy, moments accountant,  conjugate exponential family, P{\'o}lya-Gamma data augmentation.
\end{keywords}

\section{Introduction}
Bayesian inference, which reasons over the uncertainty in model parameters and latent variables given data and prior knowledge, has found widespread use in data science application domains in which privacy is essential, including text analysis \citep{blei2003latent}, medical informatics \citep{husmeier2006probabilistic}, and MOOCS \citep{piech2013tuning}.  In these applications, the goals of the analysis must be carefully balanced against the privacy concerns of the individuals whose data are being studied \citep{daries2014privacy}.  The recently proposed \emph{Differential Privacy} (DP) formalism provides a means for analyzing and controlling this trade-off, by quantifying the privacy ``cost'' of data-driven algorithms \citep{dwork2006calibrating}.   In this work, we address the challenge of performing Bayesian inference in private settings, by developing an extension of the widely used \emph{Variational Bayes} (VB) algorithm that preserves differential privacy.  We provide extensive experiments  across a variety of probabilistic models which demonstrate that our algorithm is a practical, broadly applicable, and statistically efficient method for privacy-preserving Bayesian inference.

The algorithm that we build upon, variational Bayes, is an optimisation-based approach to Bayesian inference which has origins in the closely related {\it{Expectation Maximisation}} (EM) algorithm \citep{10.2307/2984875}, although there are important differences between these methods.  Variational Bayes outputs an approximation to the full Bayesian posterior distribution, while expectation maximisation performs maximum likelihood and MAP estimation, and hence outputs a point estimate of the parameters.  Thus, VB performs fully Bayesian inference approximately, while EM does not.  It will nevertheless be convenient to frame our discussion on VB in the context of EM, following \citet{Beal_03}.  The EM algorithm seeks to learn the parameters of models with latent variables.  Since directly optimising the log likelihood of observations under such models is intractable, EM introduces a lower bound on the log likelihood by rewriting it in terms of auxiliary probability distributions over the latent variables, and using a Jensen's inequality argument.
EM proceeds by iteratively alternating between improving the bound via updating the auxiliary distributions (the E-step), and optimising the lower bound with respect to the parameters (the M-step).

Alternatively, EM can instead be understood as an instance of the \emph{variational method}, in which both the E- and M-steps are viewed as maximising the same joint objective function: a reinterpretation of the bound as a \emph{variational} lower bound which is related to a quantity known as the \emph{variational free energy} in statistical physics, and to the KL-divergence \citep{neal1998view}.
This interpretation opens the door to extensions in which simplifying assumptions are made on the optimization problem, thereby trading improved computational tractability against tightness of the bound.
Such simplifications, for instance  assuming that the auxiliary distributions are fully factorized, make feasible a fully Bayesian extension of EM, called {\it{Variational Bayesian}} EM (VBEM) \citep{Beal_03}. 
While VBEM has a somewhat similar algorithmic structure to EM, it aims to compute a fundamentally different object: an approximation to the entire posterior distribution, instead of a point estimate of the parameters.
VBEM thereby provides an optimisation-based alternative to Markov Chain Monte Carlo (MCMC) simulation methods for Bayesian inference, and as such, frequently has faster convergence properties than corresponding MCMC methods.
We collectively refer to VBEM and an intermediary method between it and EM, called \emph{Variational} EM (VEM), as \emph{variational Bayes}.\footnote{Variational EM uses simplifying assumptions on the auxiliary distribution over latent variables, as in VBEM, but still aims to find a point estimate of the parameters, as in EM.  See \secref{sec:VB} for more details on variational Bayes, and Appendix D for more details on EM.}

While the variational Bayes algorithm proves its usefulness by successfully solving many statistical problems, 
the standard form of the algorithm unfortunately cannot guarantee privacy for each individual in the dataset.
Differential Privacy (DP) \citep{dwork2006calibrating} is a formalism which can be used to establish such guarantees.
An algorithm is said to preserve differential privacy if its output is sufficiently noisy or random to obscure the participation of any single individual in the data. 
By showing that an algorithm satisfies the differential privacy criterion, we are guaranteed that an adversary cannot draw new conclusions about an individual from the output of the algorithm, by virtue of his/her participation in the dataset.
However, the injection of noise into an algorithm, in order to satisfy the DP definition, generally results in a loss of statistical efficiency, as measured by accuracy per sample in the dataset.
To design practical differentially private machine learning algorithms, the central challenge is to design a noise injection mechanism such that there is a good trade-off between privacy and statistical efficiency.

%
%
%
Iterative algorithms such as variational Bayes pose a further challenge, when developing a differentially private algorithm: each iteration corresponds to a query to the database which must be privatised, and the number of iterations required to guarantee accurate posterior estimates causes high cumulative privacy loss. To compensate for the loss, one needs to add a significantly high level of noise to the quantity of interest. 
%
%
We overcome these challenges in the context of variational Bayes by using the following key ideas:
\begin{itemize}
\item \textbf{Perturbation of the expected sufficient statistics}: Our first observation is that when models are in the {\it{Conjugate Exponential}} (CE) family, we can privatise variational posterior distributions simply by perturbing the expected sufficient statistics of the complete-data likelihood. This allows us to make effective use of the \emph{per iteration privacy budget} in each step of the algorithm.
\item \textbf{Refined composition analysis}: In order to use the privacy budget more effectively \emph{across many iterations}, we calculate the cumulative privacy cost  by using an improved composition analysis, 
the \emph{Moments Accountant} (MA) method of \citet{2016arXiv160700133A}.  This method maintains a bound on the log of the moment generating function of the privacy loss incurred by applying multiple mechanisms to the dataset, i.e. one mechanism for each iteration of a machine learning algorithm.  This allows for a higher per-iteration budget than standard methods.
\item \textbf{Privacy amplification effect from subsampling of large-scale data}: 
Processing the entire dataset in each iteration is extremely expensive in the big data setting, and is not possible in the case of streaming data, for which the size of the dataset is assumed to be infinite.  Stochastic learning algorithms provide a scalable alternative by performing parameter updates based on subsampled mini-batches of data, and this has proved to be highly advantageous for large-scale applications of variational Bayes \citep{Hoffman2013SVI}. This subsampling procedure, in fact, has a further benefit of \emph{amplifying privacy}. 
%
Our results confirm that \emph{subsampling works synergistically in concert with moments accountant composition} to make effective use of an overall privacy budget.
While there are several prior works on differentially private algorithms in stochastic learning (e.g. \citet{DBLP:conf/focs/BassilyST14, SCS13, Wangetal15, Wang16learning, WKCJN16}), the use of privacy amplification due to subsampling, together with MA composition, has not been used in the context of variational Bayes before (\citet{2016arXiv160700133A} used this approach for privacy-preserving deep learning).
\item \textbf{Data augmentation for the non-CE family models}: For widely used non-CE models with binomial likelihoods such as logistic regression, we exploit the \emph{P{\'o}lya-Gamma} data augmentation scheme \citep{PolsonScott13} to bring such models into the CE family, such that inferences in the modified model resemble our private variational Bayes algorithm as closely as possible. 
%
%
Unlike recent work which involves perturbing and clipping gradients for privacy \citep{DPVI_Nonconjugate16}, our method uses an improved composition method, and also maintains the closed-form updates for the variational posteriors and the posterior over hyper-parameters, 
and results in an algorithm which is more faithful to the standard CE variational Bayes method.
Several papers have used the P{\'o}lya-Gamma data augmentation method in order to perform Bayesian inference, either exactly via Gibbs sampling \citep{PolsonScott13}, or approximately via variational Bayes \citep{GanHCC15}. However, this augmentation technique has not previously been used in the context of differential privacy.
\end{itemize}

Taken together, these ideas result in an algorithm for privacy-preserving variational Bayesian inference that is both practical and broadly applicable. Our private VB algorithm makes effective use of the privacy budget, both per iteration and across multiple iterations, and with further improvements in the stochastic variational inference setting. Our algorithm is also general, and applies to the broad class of CE family models, as well as non-CE models with binomial likelihoods.
We present extensive empirical results demonstrating that our algorithm can preserve differential privacy while maintaining a high degree of statistical efficiency, leading to practical private Bayesian inference on a range of probabilistic models.

We organise the remainder of this paper as follows. First, we review relevant background information on differential privacy, privacy-preserving Bayesian inference, and variational Bayes in \secref{DP_VI_summary}. We then introduce  our novel general framework of private variational Bayes in \secref{general_PPVI} and illustrate how to apply that general framework to the latent Dirichlet allocation model in \secref{LDA_example}. In \secref{non-CE-family}, we introduce the P{\'o}lya-Gamma data augmentation scheme for non-CE family models, and we then illustrate how to apply our private variational Bayes algorithm to Bayesian logistic regression (\secref{bayesian_logistic_reg}) and sigmoid belief networks (\secref{sigmoid_belief_nets}).
Lastly, we summarise our paper and provide future directions in \secref{Discussion}.

\section{Background}\label{DP_VI_summary}


We begin with some background information on differential privacy, general
techniques for designing differentially private algorithms and the composition
techniques that we use.  We then provide related work on privacy-preserving
Bayesian inference, as well as the general formulation of the variational
inference algorithm. 

\subsection{Differential privacy} 

Differential Privacy (DP) is a formal
definition of the privacy properties of data analysis algorithms
\citep{Dwork14,dwork2006calibrating}. 

\begin{definition}[Differential Privacy] \label{def:dp}
A randomized algorithm $\mathcal{M}(\mathbf{X})$ is said to
be $(\epsilon,\delta)$-differentially private if
\begin{equation}
\label{eqn:DP}
Pr(\mathcal{M}(\mathbf{X}) \in \mathcal{S}) \leq \exp(\epsilon) Pr(\mathcal{M}(\mathbf{X}') \in \mathcal{S}) + \delta
\end{equation}
for all measurable subsets $\mathcal{S}$ of the range of $\mathcal{M}$ and for
all datasets $\mathbf{X}$, $\mathbf{X}'$ differing by a single entry (either by
excluding that entry or replacing it with a new entry).
\label{def:DP}
\end{definition}

 Here, an entry usually corresponds to a single individual's private value. If
$\delta = 0$, the algorithm is said to be $\epsilon$-differentially private,
and if $\delta > 0$, it is said to be {\it{approximately}} differentially
private. 

Intuitively, the definition states that the probability of any event does not
change very much when a single individual's data is modified, thereby limiting
the amount of information that the algorithm reveals about any one individual.
We observe that $\mathcal{M}$ is a randomized algorithm, and randomization is
achieved by either adding external noise, or by subsampling.  In this paper, we use the ``\emph{include/exclude}'' version of \defref{DP}, in which \emph{differing by a single entry} refers to the inclusion or exclusion of that entry in the dataset.\footnote{We use this version of differential privacy in order to make use of \citet{2016arXiv160700133A}'s analysis of the ``privacy amplification'' effect of subsampling in the specific case of MA using the Gaussian mechanism.  Privacy amplification is also possible with the ``\emph{replace-one}'' definition of DP, cf. \citet{Wang16learning}. }

\subsection{Designing differentially private algorithms}

There are several standard approaches for designing differentially-private
algorithms -- see~\citet{Dwork14} and~\cite{SarwateC13} for surveys. The
classical approach is output perturbation by \cite{dwork2006calibrating}, where
the idea is to add noise to the output of a function computed on sensitive
data. The most common form of output perturbation is the {\em{global
sensitivity method}} by \cite{dwork2006calibrating}, where the idea is to
calibrate the noise added to the global sensitivity of the function.

\subsubsection{The Global Sensitivity Mechanism}

The global sensitivity of a function $F$ of a dataset $\mathbf{X}$ is defined
as the maximum amount (over all datasets $\mathbf{X}$) by which $F$ changes
when the private value of a single individual in $\mathbf{X}$ changes.
Specifically, \[ \Delta F = \max_{|\mathbf{X} - \mathbf{X}'| = 1} |
F(\mathbf{X}) - F(\mathbf{X}') |, \] where $\mathbf{X}$ is allowed to vary over
the entire data domain, and $| F(\cdot) |$ can correspond to either the $L_1$ norm
or the $L_2$ norm of the function, depending on the noise mechanism used.

In this paper, we consider a specific form of the global sensitivity method,
called the {\em{Gaussian mechanism}} \citep{dwork2006our}, where Gaussian noise
calibrated to the global sensitivity in the $L_2$ norm is added. Specifically,
for a function $F$ with global sensitivity $\Delta F$, we output:
\[ F(\mathbf{X}) +  Z, \quad {\text{where\ }} Z \sim N(0, \sigma^2) \mbox{ , } \sigma^2 \geq 2 \log (1.25/\delta)(\Delta F)^2/\epsilon^2 \mbox{ , } \]
and where $\Delta F$ is computed using the $L_2$ norm, and is referred to as
the \emph{L2 sensitivity} of the function $F$. The privacy properties of this
method are illustrated in~\citep{Dwork14, zCDP16, CDP16}. 

\subsubsection{Other Mechanisms}

A variation of the global sensitivity method is the {\em{smoothed sensitivity
method}}~\citep{NRS07}, where the standard deviation of the added noise depends
on the dataset itself, and is less when the dataset is {\em{well-behaved}}.
Computing the smoothed sensitivity in a tractable manner is a major challenge,
and efficient computational procedures are known only for a small number of
relatively simple tasks.

A second, different approach is the {\em{exponential mechanism}}~\citep{MT07},
a generic procedure for privately solving an optimisation problem where the
objective depends on sensitive data.  The exponential mechanism outputs a
sample drawn from a density concentrated around the (non-private) optimal
value; this method too is computationally inefficient for large domains. A third
approach is {\em{objective perturbation}}~\citep{ERM}, where an optimisation
problem is perturbed by adding a (randomly drawn) term and its solution is
output; while this method applies easily to convex optimisation problems such
as those that arise in logistic regression and SVM, it is unclear how to apply
it to more complex optimisation problems that arise in Bayesian
inference. 

A final approach for designing differentially private algorithms is
sample-and-aggregate \citep{NRS07}, where the goal is to boost the amount of
privacy offered by running private algorithms on subsamples, and then
aggregating the result. In this paper, we use a combination of output
perturbation along with sample-and-aggregate. 

\subsection{Composition}

An important property of differential privacy which makes it conducive to real
applications is {\em{composition}}, which means that the privacy guarantees
decay gracefully as the same private dataset is used in multiple releases. This
property allows us to easily design private versions of iterative algorithms by
making each iteration private, and then accounting for the privacy loss
incurred by a fixed number of iterations. 

The first composition result was established by~\citet{dwork2006calibrating}
and~\cite{dwork2006our}, who showed that differential privacy composes
{\em{linearly}}; if we use differentially private algorithms $A_1, \ldots, A_k$
with privacy parameters $(\epsilon_1, \delta_1), \ldots, (\epsilon_k,
\delta_k)$ then the resulting process is $(\sum_k \epsilon_k, \sum_k
\delta_k)$-differentially private. This result was improved
by~\citet{dwork2010boosting} to provide a better rate for $(\epsilon,
\delta)$-differential privacy. \cite{kairouz2013composition} improves this even
further, and provides a characterization of optimal composition for any
differentially private algorithm.  \cite{zCDP16} uses these ideas to provide
simpler composition results for differentially mechanisms that obey certain
properties. 

\subsubsection{The Moments Accountant Method}

In this paper, we use the \emph{Moments Accountant} (MA) composition method due
to~\cite{2016arXiv160700133A} for accounting for privacy loss incurred by
successive iterations of an iterative mechanism. We choose this method as it is
tighter than~\cite{dwork2010boosting}, and applies more generally than zCDP
composition~\citep{zCDP16}.  Moreover, unlike the result
in~\cite{kairouz2013composition}, this method is tailored to specific
algorithms, and has relatively simple forms, which makes calculations easy.

The moments accountant method is based on the concept of a {\em{privacy loss
random variable}}, which allows us to consider the entire spectrum of
likelihood ratios $\frac{Pr(\mathcal{M}(\mathbf{X}) =
o)}{Pr(\mathcal{M}(\mathbf{X}') = o)}$ induced by a privacy mechanism
$\mc{M}$. Specifically, the privacy loss random variable corresponding to a
mechanism $\mathcal{M}$, datasets $\mathbf{X}$ and $\mathbf{X'}$, and an
auxiliary parameter $w$ is a random variable defined as follows:

\[ L_{\mc{M}}(\mb{X}, \mb{X'}, w) := \log \frac{Pr(\mathcal{M}(\mathbf{X}, w) = o)}{Pr(\mathcal{M}(\mathbf{X}', w) = o)}, \quad {\text{with likelihood\ }} Pr(\mathcal{M}(\mathbf{X}, w) = o), \]

where $o$ lies in the range of $\mc{M}$. Observe that if $\mc{M}$ is
$(\epsilon, 0)$-differentially private, then the absolute value of
$L_{\mc{M}}(\mb{X}, \mb{X'}, w)$ is at most $\epsilon$ with probability $1$. 

The moments accountant method exploits properties of this privacy loss random
variable to account for the privacy loss incurred by applying mechanisms
$\mc{M}_1, \ldots, \mc{M}_t$ successively to a dataset $\mb{X}$; this is done
by bounding properties of the log of the moment generating function of the
privacy loss random variable. Specifically, the log moment function
$\alpha_{\mc{M}_t}$ of a mechanism $\mc{M}_t$ is defined as:
\begin{equation}\label{eqn:logmomentfn}
\alpha_{\mc{M}_t}(\lambda) = \sup_{\mb{X}, \mb{X'}, w} \log \bbE[ \exp(\lambda  L_{\mc{M}_t}(\mb{X}, \mb{X'}, w))],
\end{equation}
where $\mb{X}$ and $\mb{X'}$ are datasets that differ in the private value of a
single person.\footnote{In this paper, we will interchangeably denote expectations as $\bbE[\cdot]$ or $\langle \cdot \rangle $.} \cite{2016arXiv160700133A} shows that if $\mc{M}$ is the combination of mechanisms $(\mc{M}_1, \ldots, \mc{M}_k)$ where each mechanism addes independent noise, then, its log moment generating function $\alpha_{\mc{M}}$ has the property that:
\begin{equation}\label{eqn:logmomentadditive}
\alpha_{\mc{M}}(\lambda) \leq \sum_{t=1}^{k} \alpha_{\mc{M}_t}(\lambda)
\end{equation}
Additionally, given a log moment function $\alpha_{\mc{M}}$, the corresponding mechanism $\mc{M}$ satisfies a range of privacy parameters $(\epsilon, \delta)$ connected by the following equation:
\begin{equation}\label{eqn:logmomenttoprivacy}
\delta = \min_{\lambda} \exp(\alpha_{\mc{M}}(\lambda) - \lambda \epsilon)
\end{equation}

These properties immediately suggest a procedure for tracking privacy loss incurred by a combination of mechanisms $\mc{M}_1, \ldots, \mc{M}_k$ on a dataset. For each mechanism $\mc{M}_t$, first compute the log moment function $\alpha_{\mc{M}_t}$; for simple mechanisms such as the Gaussian mechanism this can be done by simple algebra. Next, compute $\alpha_{\mc{M}}$ for the combination $\mc{M} = (\mc{M}_1, \ldots, \mc{M}_k)$ from~(\ref{eqn:logmomentadditive}), and finally, recover the privacy parameters of $\mc{M}$ using~(\ref{eqn:logmomenttoprivacy}) by either finding the best $\epsilon$ for a target $\delta$ or the best $\delta$ for a target $\epsilon$. In some special cases such as composition of $k$ Gaussian mechanisms, the log moment functions can be calculated in closed form; the more common case is when closed forms are not available, and then a grid search may be performed over $\lambda$. We observe that any $(\epsilon, \delta)$ obtained as a solution to~(\ref{eqn:logmomenttoprivacy}) via grid search are still valid privacy parameters, although they may be suboptimal if the grid is too coarse.

\subsection{Privacy-preserving Bayesian inference}\label{related_work}

Privacy-preserving Bayesian inference is a new research area which is currently receiving a lot of attention. \citet{dimitrakakis2014robust} showed that Bayesian posterior sampling is automatically differentially private, under a mild sensitivity condition on the log likelihood.  This result was independently discovered by \citet{Wangetal15}, who also showed that the Stochastic Gradient Langevin Dynamics (SGLD) Bayesian inference algorithm \citep{welling2011bayesian} automatically satisfies approximate differential privacy, due to the use of Gaussian noise in the updates.

As an alternative to obtaining privacy ``for free'' from posterior sampling, \citet{FGWC16} studied a Laplace mechanism approach for exponential family models, and proved that it is asymptotically efficient, unlike the former approach. Independently of this work, \citet{zhang2016differential} proposed an equivalent Laplace mechanism method for private Bayesian inference in the special case of beta-Bernoulli systems, including graphical models constructed based on these systems. \citet{FGWC16} further analyzed privacy-preserving MCMC algorithms, via both exponential and Laplace mechanism approaches.

In terms of approximate Bayesian inference, \citet{DPVI_Nonconjugate16}
recently considered privacy-preserving variational Bayes via perturbing and clipping the gradients of the variational lower bound. 
However, this work focuses its experiments on logistic regression, a model that does not have latent variables. 
Given that most latent variable models consist of at least as many latent variables as the number of datapoints, the long vector of gradients (the concatenation of the gradients with respect to the latent variables; and with respect to the model parameters) in such cases is expected to typically require excessive amounts of additive noise.
Furthermore, our approach, using the data augmentation scheme (see Sec. \ref{non-CE-family}) and moment perturbation, yields closed-form posterior updates (posterior distributions both for latent variables and model parameters) that are closer to the spirit of the original variational Bayes method, for both CE and non-CE models, as well as an improved composition analysis using moments accountant. 

In recent work, \citet{BartheFGAGHS16} designed a {\it{probabilistic programming}} language for designing privacy-preserving Bayesian machine learning algorithms, with privacy achieved via input or output perturbation, using standard mechanisms.

Lastly, although it is not a fully Bayesian method, it is worth noting the differentially private expectation maximisation algorithm developed by \citet{DPEM16}, which also involves perturbing the expected sufficient statistics for the complete-data likelihood. The major difference between our and their work is that EM is not (fully) Bayesian, i.e., EM outputs the {\it{point estimates}} of the model parameters; while VB outputs the posterior distributions (or those quantities that are necessary to do Bayesian predictions, e.g., expected natural parameters and expected sufficient statistics).  Furthermore, \citet{DPEM16} deals with only CE family models for obtaining the closed-form MAP estimates of the parameters; while our approach encompasses both CE and non-CE family models. Lastly, \citet{DPEM16} demonstrated their method on small- to medium-sized datasets, which do not require stochastic learning; while our method takes into account the scenario of stochastic learning which is essential in the era of big data \citep{Hoffman2013SVI}.

\subsection{Variational Bayes}
\label{sec:VB}
\emph{Variational inference} is the class of techniques which solve inference problems in probabilistic models using variational methods \citep{jordan1999introduction, Wainwright08}.  The general idea of variational methods is to cast a quantity of interest as an optimisation problem.  By relaxing the problem in some way, we can replace the original intractable problem with one that we can solve efficiently.\footnote{Note that the ``variational'' terminology comes from the calculus of variations, which is concerned with finding optima of functionals.  This pertains to probabilistic inference, since distributions are functions, and we aim to optimise over the space of possible distributions. However, it is typically not necessary to use the calculus of variations when deriving variational inference algorithms in practice.}

The application of variational inference to finding a Bayesian posterior distribution is called \emph{Variational Bayes} (\emph{VB}).  
Our discussion is focused on the Variational Bayesian EM (VBEM) algorithm variant.
The goal of the algorithm is to compute an approximation $q$ to the posterior distribution over latent variables and model parameters for models where exact posterior inference is intractable.
This should be contrasted to VEM and EM, which aim to compute a point estimate of the parameters.
VEM and EM can both be understood as special cases, in which the set of $q$ distributions is constrained such that the approximate posterior is a Dirac delta function \citep{Beal_03}.  We therefore include them within the definition of VB. See \citet{blei2017variational} for a recent review on variational Bayes, \citet{Beal_03} for more detailed derivations, and see Appendix D for more information on the relationship between EM and VBEM.

\paragraph{High-level derivation of VB} 
Consider a generative model that produces a dataset $\Dat = \{ \Dat_n \}_{n=1}^N$ consisting of $N$ independent identically distributed ($i.i.d.$) items ($\Dat_n$ is the $n$th input/output pair $\{\vx_n, y_n \} $ for supervised learning, and $\Dat_n$ is the $n$th vector output $\vy_n$ for unsupervised learning), generated using a set of latent variables $\vl = \{ \vl_n \}_{n=1}^N$. 
The generative model provides $p(\Dat_n| \vl_n,\vm)$, where $\vm$ are the model parameters.
We also consider the prior distribution over the model parameters $p(\vm)$ and the prior distribution over the latent variables $p(\vl)$. 

Variational Bayes recasts the task of approximating the posterior $p(\vl,\vm|\Dat)$ as an optimisation problem: making the approximating distribution $q(\vl,\vm)$, which is called the \emph{variational distribution}, as similar as possible to the posterior, by minimising some distance (or divergence) between them.  The terminology \emph{VB} is often assumed to refer to the standard case,  in which the divergence to minimise is the KL-divergence from $p(\vl,\vm|\Dat)$ to $q(\vl,\vm)$,
\begin{align}
D_{KL}(q(\vl,\vm)\|p(\vl,\vm|\Dat)) &= \bbE_q \Big[\log \frac{ q(\vl,\vm)}{ p(\vl,\vm|\Dat)}\Big] \nonumber \\
&= \bbE_q[\log q(\vl,\vm)] - \bbE_q [\log{p(\mathbf{\vl,\vm}|\Dat)}] \nonumber\\
&= \bbE_q[\log q(\vl,\vm)] - \bbE_q [\log{p(\mathbf{\vl,\vm},\Dat)}] + \log p(\Dat) \label{eq:VB_KL}\mbox{ .}
\end{align}
The $\arg \min$ of \eqref{VB_KL} with respect to $q(\vl,\vm)$ does not depend on the constant $\log p(\Dat)$.  Minimising it is therefore equivalent to maximizing
\begin{equation}
  \mathcal{L}(q) \triangleq  \bbE_q [\log{p(\vl,\vm,\Dat)}] - \bbE_q[\log q(\vl,\vm)] = \bbE_q [\log{p(\vl,\vm,\Dat)}] + H(q) \mbox{ ,} \label{eq:VB_ELBO_fromKL}
\end{equation}
where $H(q)$ is the entropy (or differential entropy) of $q(\vl,\vm)$.  The entropy of $q(\vl,\vm)$ rewards simplicity, while $\bbE_q [\log{p(\vl,\vm,\Dat)}]$, the expected value of the complete data log-likelihood under the variational distribution, rewards accurately fitting to the data.  Since the KL-divergence is 0 if and only if the two distributions are the same, maximizing $\mathcal{L}(q)$ will result in $q(\vl,\vm) = p(\vl,\vm|\Dat)$ when the optimisation problem is unconstrained.  In practice, however, we restrict $q(\vl,\vm)$ to a tractable subset of possible distributions.  A common choice for the tractable subset is the set of fully factorized distributions $q(\vl,\vm) = q(\vm) \prod_{n=1}^N q(\vl_n)$, in which case the method is referred to as \emph{mean-field variational Bayes}.

We can alternatively derive $\mathcal{L}(q)$ as a lower bound on the log of the marginal probability of the data $p(\Dat)$ (the \emph{evidence}),
\begin{align}
  \log p(\Dat) &= \log \Big (\int d\vl \; d\vm \;  p(\vl,\vm,\Dat) \Big )
   = \log \Big (\int d\vl \; d\vm \;  p(\vl,\vm,\Dat) \frac{q(\vl,\vm)}{q(\vl,\vm)} \Big ) \nonumber \\
   &= \log \Big (\bbE_q \Big [\frac{p(\vl,\vm,\Dat)}{q(\vl,\vm)}\Big ] \Big)
   \geq \bbE_q \Big [\log p(\vl,\vm,\Dat) - \log q(\vl,\vm)\Big ] \label{eq:VB_ELBO}
   = \mathcal{L}(q) \mbox{ ,}
\end{align}
where we have made use of Jensen's inequality.  Due to \eqref{VB_ELBO}, $\mathcal{L}(q)$ is sometimes referred to as a variational lower bound, and in particular, the \emph{Evidence Lower Bound} (ELBO), since it is a lower bound on the log of the model evidence $p(\Dat)$.  The VBEM algorithm maximises $\mathcal{L}(q)$ via coordinate ascent over parameters encoding $q(\vl, \vm)$, called the variational parameters.  The E-step optimises the variational parameters pertaining to the latent variables $\vl$, and the M-step optimises the variational parameters pertaining to the model parameters $\vm$.    Under certain assumptions, these updates have a certain form, as we will describe below.  These updates are iterated until convergence.

\paragraph{VB for CE models}
VB simplifies to a two-step procedure when the model falls into the Conjugate-Exponential (CE) class of models, which satisfy two conditions \citep{Beal_03}:
\begin{align}
&(1)\mbox{ The complete-data likelihood is in the exponential family}: \nonumber \\
& \qquad p(\Dat_n, \vl_n| \vm) = g(\vm) f(\Dat_n, \vl_n) \exp(\vn(\vm)\trp \vs(\Dat_n, \vl_n)), \\
&(2)\mbox{ The prior over $\vm$ is conjugate to the complete-data likelihood}: \nonumber \\
& \qquad p(\vm|\tau, \vnu) = h(\tau, \vnu) g(\vm)^{\tau} \exp(\vnu\trp\vn(\vm)).
\end{align} where natural parameters and sufficient statistics of the complete-data likelihood are denoted by  $\vn(\vm)$ and $\vs(\Dat_n, \vl_n)$, respectively, and $g,f,h$ are some known functions. The hyperparameters are denoted by $\tau$ (a scalar) and $\vnu$ (a vector).

A large class of models fall in the CE family. Examples include linear dynamical systems and switching models; Gaussian mixtures; factor analysis and probabilistic PCA; Hidden Markov Models (HMM) and factorial HMMs; and discrete-variable belief networks. The models that are widely used but not in the CE family include: Markov Random Fields (MRFs) and Boltzmann machines; logistic regression; sigmoid belief networks; and Independent Component Analysis (ICA). We illustrate how best to bring such models into the CE family in a later section.

The VB algorithm for a CE family model optimises the lower bound on the model log marginal likelihood given by \eqref{VB_ELBO} (the ELBO),
\begin{equation}\label{eq:var_lbd}
\mathcal{L}(q(\vl) q(\vm)) = \int d\vm \; d\vl \; q(\vl) q(\vm) \log \frac{p(\vl, \Dat, \vm)}{q(\vl) q(\vm)}, 
\end{equation}
where we assume that the joint approximate posterior distribution over the latent variables and model parameters $q(\vl, \vm)$ is factorised via the mean-field assumption as 
\begin{align}
q(\vl, \vm) = q(\vl) q(\vm) = q(\vm) \prod_{n=1}^N q(\vl_n),
\end{align}
and that each of the variational distributions  also has the form of an exponential family distribution.
Computing the derivatives of the variational lower bound in  \eqref{var_lbd} with respect to each of these variational distributions and setting them to zero yield the following two-step procedure. 
\begin{align}\label{eq:VB}
&(1) \mbox{ First, given expected natural parameters $\bar{\vn}$, the E-step computes:}    \nonumber \\
& \qquad  q(\vl) = \prod_{n=1}^N q(\vl_n) \propto  \prod_{n=1}^N f(\Dat_n, \vl_n) \exp(\bar{\vn}\trp \vs(\Dat_n, \vl_n)) = \prod_{n=1}^N p(\vl_n|\Dat_n, \bar{\vn}) .  \\
& \qquad \mbox{Using $q(\vl)$, it outputs expected sufficient statistics,  the  expectation  of  ${\vs}(\Dat_n, \vl_n)$}\nonumber \\
& \qquad \mbox{with probability density $q(\vl_n)$ : } \bar{\vs}(\Dat) = \tfrac{1}{N} \sum_{n=1}^N \langle {\vs}(\Dat_n, \vl_n) \rangle_{q(\vl_n)}. \nonumber \\
&(2)\mbox{ Second, given expected sufficient statistics $\bar{\vs}(\Dat)$, the M-step computes:}  \nonumber \\
&  \qquad q(\vm) =  h(\tilde{\tau}, \tilde{\vnu}) g(\vm)^{\tilde{\tau}} \exp(\tilde{\vnu} \trp \vn(\vm)), \mbox{ where } \tilde{\tau} = \tau + N, \; \tilde{\vnu} = \vnu + N \bar{\vs}(\Dat).\\
& \qquad \mbox{Using $q(\vm)$, it outputs expected natural parameters $\bar{\vn}= \langle \vn(\vm) \rangle_{q(\vm)}$}.\nonumber
\end{align} 

\paragraph{Stochastic VB for CE models}

The VB update introduced in \eqref{VB} is inefficient for large data sets because we should optimise
the variational posterior over the latent variables corresponding to each data point before re-estimating the  variational posterior over the parameters. For more efficient learning, we adopt stochastic variational inference, which uses stochastic optimisation to fit the variational distribution over the parameters. We repeatedly subsample the data to form noisy estimates of the natural gradient
of the variational lower bound, and we follow these estimates with a decreasing step-size $\rho_t$, as in \citet{Hoffman2013SVI}.\footnote{When optimising over a probability distribution, the Euclidean distance between two parameter vectors is often a poor measure of the dissimilarity of the distributions. The natural gradient of a function accounts for the information geometry of its parameter space, using a Riemannian metric to adjust the direction of the traditional gradient, which results in a faster convergence than the traditional gradient \citep{Hoffman2013SVI}. } 
The stochastic variational Bayes algorithm 
is summarised in Algorithm \ref{algo:SVI_for_CE}. 
\begin{algorithm}[h]
\caption{(Stochastic) Variational Bayes  for CE family distributions}
\label{algo:SVI_for_CE}
\begin{algorithmic}
\REQUIRE Data $\Dat$. Define $\rho_{t} = (\tau_0 + t)^{-\kappa}$ and mini-batch size $S$.
\ENSURE Expected natural parameters $\bar{\vn}$ and expected sufficient statistics $\bar{\vs}$.
\FOR{$t = 1, \ldots, J $}
\STATE {\it{\textbf{(1) E-step}}}: Given the expected natural parameters $\bar{\vn}$, compute $q(\vl_{n})$ 
for $n=1,\ldots,S$. Output the expected sufficient statistics $\bar{\vs} = \tfrac{1}{S} \sum_{n=1}^S \langle \vs( \Dat_n, \vl_n) \rangle_{q(\vl_{n})}$.
\STATE {\it{\textbf{(2) M-step}}}: Given $ \bar{\vs} $,
compute $q(\vm)$ by $\tilde{\vnu}^{(t)} = \vnu + N \bar{\vs} $.
Set $\tilde{\vnu}^{(t)} \mapsfrom (1-\rho_t)\tilde{\vnu}^{(t-1)} + \rho_t \tilde{\vnu}^{(t)} $. 
Output the expected natural parameters $\bar{\vn} = \langle \vm \rangle_{q(\vm)}$.
\ENDFOR
\end{algorithmic}
\end{algorithm}

\section{Variational Bayes In Private Settings (VIPS) for the CE family}\label{general_PPVI} 


To create an extension of variational Bayes which preserves differential privacy, we need to inject noise into the algorithm.
The design choices for the noise injection procedure must be carefully made, as they can strongly affect the statistical efficiency of the algorithm, in terms of its accuracy versus the number of samples in the dataset.
We start by introducing our problem setup.
 
\subsection{Problem setup}
A naive way to privatise the VB algorithm is by perturbing both $q(\vl)$ and $q(\vm)$. 
Unfortunately, this is impractical, due to the excessive amounts of additive noise (recall: we have as many latent variables as the number of datapoints). 
We propose to perturb the expected sufficient statistics {\it{only}}. What follows next explains why this makes sense. 

\begin{figure}[t]
\centering
\includegraphics[width=1\textwidth]{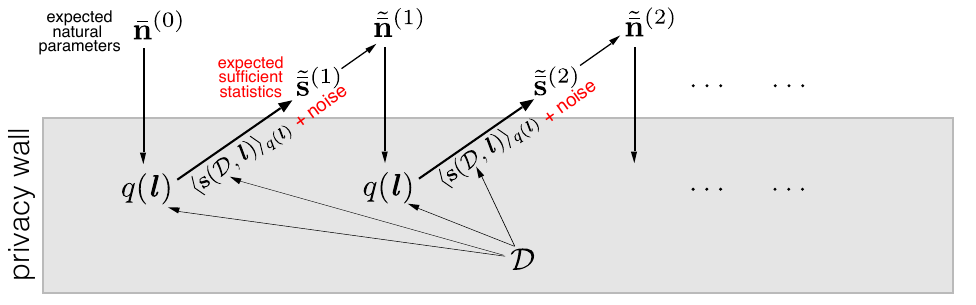}
  \caption{Schematic of VIPS. 
Given the initial expected natural parameters $\bar{\vn}^{(0)}$, we compute the variational posterior over the latent variables $q(\vl)$. Since $q(\vl)$ is a function of not only the expected natural parameters but also the data $\Dat$, we compute $q(\vl)$ behind the privacy wall. Using  $q(\vl)$, we then compute the expected sufficient statistics. 
Note that we neither perturb nor output $q(\vl)$ itself. Instead, when we noise up the expected sufficient statistics before outputting,  we add noise to each coordinate of the expected sufficient statistics in order to  compensate the maximum difference in $\langle \vs_l(\Dat_j, \vl_j)\rangle_{q(\vl_j)}$ caused by both $\Dat_j$ and $q(\vl_j)$. 
In the M-step, we compute the variational posterior over the parameters $q(\vm)$ using the perturbed expected sufficient statistics $\tilde{\bar{\vs}}^{(1)}$.  Using $q(\vm)$, we compute the expected natural parameters $\tilde{\bar{\vn}}^{(1)}$, which is already perturbed since it is a function of $\tilde{\bar{\vs}}^{(1)}$. We continue performing these two steps until convergence.} 
\label{fig:VIPS}
\end{figure}

While the VB algorithm is being run, the places where the algorithm needs to look at the data are (1) when computing the variational posterior over the latent variables $q(\vl)$; and (2) when computing the expected sufficient statistics  $\bar{\vs}(\Dat)$ given $q(\vl)$ in the E-step. 
In our proposed approach, we compute $q(\vl)$ behind the {\it{privacy wall}} (see below), and compute the expected sufficient statistics using $q(\vl)$, as shown in \figref{VIPS}. Before outputting the expected sufficient statistics, we perturb each coordinate of the expected sufficient statistics to  compensate the maximum difference in $\langle \vs_l(\Dat_j, \vl_j)\rangle_{q(\vl_j)}$ caused by both $\Dat_j$ and $q(\vl_j)$. 
The perturbed expected sufficient statistics then dictate the expected natural parameters in the M-step. Hence we do not need an additional step to add noise to $q(\vm)$. 

The reason we neither perturb nor output $q(\vl)$ for training data is that we do not need $q(\vl)$ itself most of the time. For instance, when computing the predictive probability
for test datapoints $\Dat_{tst}$, we need to perform the $E$-step to obtain  the variational posterior for the test data $q(\vl_{tst})$, which is a function of the test data and the expected natural parameters $\bar{\vn}$, given as
\begin{equation}\label{eq:pred_prob}
p(\Dat_{tst}|\Dat) = \int d\vm d\vl_{tst} \; p(\Dat_{tst}|\vl_{tst}, \vm) \; q(\vl_{tst}; \Dat_{tst}, \bar{\vn}) \; q(\vm; \tilde{\vnu}), 
\end{equation} where the dependence on the training data $\Dat$ is implicit in the approximate posteriors $q(\vm)$ through $\tilde{\vnu}$; and the expected natural parameters $\bar{\vn}$. Hence, outputting the perturbed sufficient statistics and the expected natural parameters suffice for protecting the privacy of individuals in the training data. Furthermore, the M-step can be performed based on the (privatised) output of the E-step, without querying the data again, so we do not need to add any further noise to the M-step to ensure privacy, due to the fact that differential privacy is immune to data-independent post-processing. To sum up, we provide our problem setup as below.
\begin{enumerate}
\item Privacy wall: We assume that the sensitive training dataset is only accessible through a sanitising interface which we call a {\it{privacy wall}}. The training data stay behind the privacy wall, and adversaries have access to the outputs of our algorithm only, i.e., no direct access to the training data, although they may have further prior knowledge on some of the individuals that are included in the training data. 
\item Training phase: Our differentially private VB algorithm releases the perturbed expected natural parameters and perturbed expected sufficient statistics in every iteration. Every release of a perturbed parameter based on the training data triggers an update in the log moment functions (see Sec 3.2 on how these are updated). At the end of the training phase, the final privacy parameters are calculated based on~(\ref{eqn:logmomenttoprivacy}).

\item Test phase: Test data are public (or belong to users), i.e., outside the privacy wall. Bayesian prediction on the test data is possible using the released expected natural parameters and expected sufficient statistics (given as \eqref{pred_prob}). Note that we do not consider protecting the privacy of the individuals in the test data.
\end{enumerate}


%

\subsection{How to compute the log moment functions?} 

%

Recall that to use the moments accountant method, we need to compute the log moment functions $\alpha_{\mc{M}_t}$ for each individual iteration $t$. An iteration $t$ of VIPS randomly subsamples a $\nu$ fraction of the dataset and uses it to compute the sufficient statistics which is then perturbed via the Gaussian mechanism with variance $\sigma^2 \mb{I}$. How the log moment function is computed depends on the sensitivity of the sufficient statistics as well as the underlying mechanism; we next discuss how each of these aspects.

\paragraph{Sensitivity analysis}

Suppose there are two neighbouring datasets $\Dat$ and $\Dat'$, where $\Dat'$ has one entry difference compared to $\Dat$ (i.e., by removing one entry from $\Dat$).

We denote the vector of expected sufficient statistics by 
$\bar{\vs}(\Dat) = \frac{1}{N}\sum_{n=1}^N \bar{\vs}(\Dat _n) =  [  M_1, \cdots, M_L ], $
where each expected sufficient statistic $M_l$ is given by 
\begin{equation}
M_l = \tfrac{1}{N}\sum_{i=1}^N \langle   \vs_l(\Dat_i, \vl_i) \rangle_{q(\vl_i)} \mbox{ .}
\end{equation} 

When computing the sensitivity of the expected sufficient statistics,
we assume, without loss of generality, the last entry removed from $\Dat$ maximises the difference in the expected sufficient statistics run on the two datasets $\Dat$ and $\Dat'$. 
Under this assumption and the $i.i.d.$ assumption on the likelihood, given the current expected natural parameters $\bar{\vn}$, all $q(\vl_i)$ and $ \vs_l(\Dat_i, \vl_i)$ for $i \in \{1, \ldots, N-1\}$ evaluated on the dataset $\Dat$ are the same as  $q(\vl'_i)$ and  $\vs_l(\Dat'_i, \vl'_i)$ evaluated on the dataset $\Dat'$. Hence, the sensitivity  is given by
 \begin{eqnarray}\label{eq:gen_sen}
 \Delta M_l &=& \max_{|\Dat \setminus\Dat'|_1 = 1} | M_l(\Dat) - M_l(\Dat')|, \nonumber \\
&=& \max_{|\Dat \setminus\Dat'|_1 = 1} |\tfrac{1}{N}\sum_{i=1}^N\mathbb{E}_{q(\vl_i)}  \vs_l(\Dat_i, \vl_i)-  \tfrac{1}{N-1} \sum_{i=1}^{N-1} \mathbb{E}_{q(\vl'_i)}  \vs_l(\Dat'_i, \vl'_i)|, \nonumber \\
&=& \max_{|\Dat \setminus\Dat'|_1 = 1} |\tfrac{1}{N}\sum_{i=1}^N\mathbb{E}_{q(\vl_i)}  \vs_l(\Dat_i, \vl_i)-  \tfrac{1}{N-1} \sum_{i=1}^{N-1} \mathbb{E}_{q(\vl_i)}  \vs_l(\Dat_i, \vl_i)|, \nonumber \\
&& \qquad \qquad \mbox{since $\mathbb{E}_{q(\vl_i)}  \vs_l(\Dat_i, \vl_i) = \mathbb{E}_{q(\vl'_i)}  \vs_l(\Dat'_i, \vl'_i)$ for ${i=\{1, \ldots, N-1\}}$}, \nonumber \\
&\leq& \max_{|\Dat \setminus\Dat'|_1 = 1} |\tfrac{N-1}{N} \tfrac{1}{N-1} \sum_{i=1}^{N-1}\mathbb{E}_{q(\vl_i)}  \vs_l(\Dat_i, \vl_i)
+ \tfrac{1}{N} \mathbb{E}_{q(\vl_N)}  \vs_l(\Dat_N, \vl_N)
-  \tfrac{N}{N}\tfrac{1}{N-1} \sum_{i=1}^{N-1}\mathbb{E}_{q(\vl_i)}  \vs_l(\Dat_i, \vl_i)| \nonumber \\
%
%
&\leq& \max_{|\Dat \setminus\Dat'|_1 = 1} | - \tfrac{1}{N}\tfrac{1}{N-1}\sum_{i=1}^{N-1}\mathbb{E}_{q(\vl_i)}  \vs_l(\Dat_i, \vl_i) +  \tfrac{1}{N} \mathbb{E}_{q(\vl_N)}  \vs_l(\Dat_N, \vl_N)|, \nonumber \\
&\leq&\max_{\Dat_N, q(\vl_N)} \tfrac{2}{N} |\mathbb{E}_{q(\vl_N)}  \vs_l(\Dat_N, \vl_N)|,
\end{eqnarray}
where the last line is because the average (over the $N-1$ terms) expected sufficient statistics is less than equal to the maximum expected sufficient statistics (recall: we assumed that the last entry maximises the difference in the expected sufficient statistics), i.e., 
\begin{eqnarray}
| \tfrac{1}{N-1}\sum_{i=1}^{N-1}\mathbb{E}_{q(\vl_i)}  \vs_l(\Dat_i, \vl_i)| 
\leq \tfrac{1}{N-1}\sum_{i=1}^{N-1}| \mathbb{E}_{q(\vl_i)}  \vs_l(\Dat_i, \vl_i)|
\leq |\mathbb{E}_{q(\vl_N)}  \vs_l(\Dat_N, \vl_N)|,
\end{eqnarray} and due to the triangle inequality. 

As in many existing works (e.g., \citep{ERM, Kifer12privateconvex}, among many others), we also assume that the dataset is pre-processed such that the $L_2$ norm of any $\Dat_i$ is less than $1$.
Furthermore, 
we choose $q(\vl_i)$ such that its support is bounded.
Under these conditions, each coordinate of the expected sufficient statistics $ \langle \vs_l (\Dat_i, \vl_i) \rangle_{q(\vl_i)}$ has  limited sensitivity.
We will add noise to each coordinate of the expected sufficient statistics to compensate this bounded maximum change in the E-step.

\paragraph{Log Moment Function of the Gaussian Mechanism with Subsampled data}

Observe that iteration $t$ of our algorithm subsamples a $\nu$ fraction of the dataset, computes the sufficient statistics based on this subsample, and perturbs it using the Gaussian mechanism with variance $\sigma^2 I_d$. To simplify the privacy calculations, we assume that each example in the dataset is included in a minibatch according to an independent coin flip with probability $\nu$. This differs slightly from the standard assumption for stochastic learning algorithms (e.g., stochastic gradient descent), in which a fixed minibatch size $S$ is typically used in each iteration.  However, following \citet{2016arXiv160700133A}, for simplicity of analysis, we will also assume that the instances are included independently with probability $\nu = \frac{S}{N}$; and for ease of implementation, we will use minibatches with fixed size $S$ in our experiments. 

From Proposition 1.6 in~\cite{zCDP16} along with simple algebra, the log moment function of the Gaussian Mechanism $\mc{M}$ applied to a query with $L_2$-sensitivity $\Delta$ is $\alpha_{\mc{M}}(\lambda) = \frac{\lambda (\lambda + 1) \Delta^2}{2 \sigma^2}$. To compute the log moment function for the subsampled Gaussian Mechanism, we follow~\cite{2016arXiv160700133A}. Let $\beta_0$ and $\beta_1$ be the densities $\mc{N}(0, (\sigma/\Delta)^2)$ and $\mc{N}(1, (\sigma/\Delta)^2)$, and let $\beta = (1 - \nu) \beta_0 + \nu \beta_1$ be the mixture density; then, the log moment function at $\lambda$ is $\max \log (E_1, E_2)$ where $E_1 = \bbE_{z \sim \beta_0} [ (\beta_0(z)/\beta(z))^{\lambda}]$ and $E_2 = \bbE_{z \sim \beta} [ (\beta(z)/\beta_0(z))^{\lambda}]$. $E_1$ and $E_2$ can be numerically calculated for any $\lambda$, and we maintain the log moments over a grid of $\lambda$ values. 

Note that our algorithms are run for a prespecified number of iterations, and with a prespecified $\sigma$; this ensures that the moments accountant analysis is correct, and we do not need an a data-dependent adaptive analysis such as in~\citet{rogers2016privacy}. 


Also, note that when using the subsampled data per iteration,  the sensitivity analysis has to be modified as now the query is evaluated on a smaller dataset. Hence, the $1/N$ factor has to be changed to $1/S$ in \eqref{gen_sen}:
\begin{align}
 \Delta M_l &\leq\max_{\Dat_S, q(\vl_S)} \tfrac{2}{S} |\mathbb{E}_{q(\vl_S)}  \vs_l(\Dat_S, \vl_S)| \mbox{ .}
\end{align}

Algorithm \ref{algo:VIPS_for_CE} summarizes our VIPS algorithm that performs differentially private stochastic variational Bayes for CE family models. 
\begin{algorithm}[h]
\caption{Private VIPS for CE family distributions}
\label{algo:VIPS_for_CE}
\begin{algorithmic}
\REQUIRE Data $\Dat$. Define $\rho_{t} = (\tau_0 + t)^{-\kappa}$, noise variance $\sigma^2$, mini-batch size $S$, and maximum iterations $J$.
\ENSURE Perturb expected natural parameters $\tilde{\bar{\vn}}$ and expected sufficient statistics $\tilde{\bar{\vs}}$.
\STATE Compute the L2-sensitivity $\Delta$ of the expected sufficient statistics. 
\FOR{$t = 1, \ldots, J $}
\STATE {\it{\textbf{(1) E-step}}}: Given the expected natural parameters $\bar{\vn}$, compute $q(\vl_{n})$ 
for $n=1,\ldots,S$.  Perturb each coordinate of $\bar{\vs} = \tfrac{1}{S} \sum_{n=1}^S \langle \vs( \Dat_n, \vl_n) \rangle_{q(\vl_{n})}$ by adding $\mc{N}(0, \sigma^2\Delta^2 I)$ noise, and output $ \tilde{\bar{\vs}} $. Update the log moment functions. 
\STATE {\it{\textbf{(2) M-step}}}: Given $ \tilde{\bar{\vs}} $,
compute $q(\vm)$ by $\tilde{\vnu}^{(t)} = \vnu + N \tilde{\bar{\vs}} $.
Set $\tilde{\vnu}^{(t)} \mapsfrom (1-\rho_t)\tilde{\vnu}^{(t-1)} + \rho_t \tilde{\vnu}^{(t)} $. 
Output the expected natural parameters $\tilde{\bar{\vn}} = \langle \vm \rangle_{q(\vm)}$.
\ENDFOR
\end{algorithmic}
\end{algorithm}

\section{VIPS for latent Dirichlet allocation}\label{LDA_example}
Here, we illustrate how to use the general framework of VIPS for CE family in the example of Latent Dirichlet Allocation (LDA).     

\subsection{Model specifics in LDA}

The most widely used topic model is Latent Dirichlet Allocation (LDA) \citep{blei2003latent}. Its generative process is given by
\begin{itemize}
\item Draw topics $\vbeta_k \sim $ Dirichlet $(\eta \vone_{V})$, for $k=\{1,\ldots, K\}$, where $\eta$ is a scalar hyperarameter.
\item For each document $d \in \{ 1, \ldots, D \}$
\begin{itemize}
\item Draw topic proportions $ \vtheta_d \sim$ Dirichlet $(\alpha \vone_{K})$, where $\alpha$ is a scalar hyperarameter.
\item For each word $n \in \{ 1, \ldots, N \}$
\begin{itemize}
\item Draw topic assignments $\vz_{dn} \sim$ Discrete$(\vtheta_d)$
\item Draw word $\vw_{dn} \sim$ Discrete$(\vbeta_{\vz_{dn}})$
\end{itemize}
\end{itemize}
\end{itemize} 
where each observed word is represented by an indicator vector $\vw_{dn}$ ($n$th word in the $d$th document) of length $V$, and where $V$ is the number of terms in a fixed vocabulary set. The topic assignment latent variable $\vz_{dn}$ is also an indicator vector of length $K$, where $K$ is the number of topics.

The LDA model falls in the CE family, viewing $\vz_{d, 1:N}$ and  $\vtheta_d $ as two types of latent variables: $\vl_d = \{ \vz_{d, 1:N}, \vtheta_d \}$, and $\vbeta$ as model parameters $\vm = \vbeta$. The conditions for CE are satisfied: (1) the complete-data likelihood per document is in exponential family: 
\begin{align}
p(\vw_{d, 1:N}, \vz_{d, 1:N}, \vtheta_d| \vbeta) \propto  f(\Dat_d, \vz_{d, 1:N}, \vtheta_d) \exp ( \sum_{n}\sum_{k} [\log \vbeta_k] \trp [\vz_{dn}^k \vw_{dn} ] ),
\end{align}
where   $f(\Dat_d, \vz_{d, 1:N}, \vtheta_d) \propto\exp([\alpha \vone_K] \trp [\log \vtheta_d] + \sum_{n}\sum_{k}\vz_{dn}^k \log \vtheta_{d}^k) $; and
(2) we have a conjugate prior over $\vbeta_k$:  
\begin{align}
p(\vbeta_k|\eta \vone_{V}) \propto \exp([\eta \vone_{V}]\trp [\log \vbeta_k]), 
\end{align}
 for  $k=\{1, \ldots, K\}$.
For simplicity, we assume hyperparameters $\alpha$ and $\eta$ are set manually.

Under the LDA model, we assume the variational posteriors are given by 
\begin{itemize}
\item Discrete : $q(\vz_{dn}^k|\vphi_{dn}^k) \propto  \exp(\vz_{dn}^k \log \vphi_{dn}^k)$, with variational parameters for capturing the posterior topic assignment, 
\begin{equation}\label{eq:topic_post}
\vphi_{dn}^k \propto \exp(\langle \log \vbeta_k \rangle_{q(\vbeta_k)} \trp \vw_{dn} + \langle \log \vtheta_d^k \rangle_{q(\vtheta_d)}).
\end{equation}
\item Dirichlet : $q(\vtheta_d | \vgamma_d) \propto \exp( \vgamma_d \trp \log \vtheta_d), \mbox{where  } \vgamma_d =  \alpha \vone_K + \sum_{n=1}^N \langle \vz_{dn} \rangle_{q(\vz_{dn})}$, 
\end{itemize} where these two distributions are computed iteratively in the E-step behind the privacy wall.  
 The expected sufficient statistics are $\bar{\vs}_k = \tfrac{1}{D}\sum_{d} \sum_{n} \langle \vz_{dn}^k \rangle_{q(\vz_{dn})} \vw_{dn} = \tfrac{1}{D}\sum_{d} \sum_{n} \vphi_{dn}^k \vw_{dn}$ due to \eqref{topic_post}.
Then, in the M-step, we compute the posterior 
\begin{itemize}
\item Dirichlet : $q(\vbeta_k|\vlambda_k  ) \propto \exp( \vlambda_k \trp \log \vbeta_k), 
 \mbox{ where }  \vlambda_k = \eta \vone_V + \sum_{d} \sum_{n} \langle \vz_{dn}^k \rangle_{q(\vz_{dn})} \vw_{dn} $.
\end{itemize}

\subsection{VIPS for LDA}
We follow the general framework of VIPS for differentially private LDA, with the addition of several LDA-specific heuristics which are important for good performance, described below.
First, while each document originally has a different document length $N_d$, in order to bound the sensitivity, and to ensure that the signal-to-noise ratio remains reasonable for very short documents, we preprocess all documents to have the same fixed length $N$.  We accomplish this by sampling $N$ words with replacement from each document's bag of words.  In our experiments, we use $N=500$.

To perturb the expected sufficient statistics, which is a matrix of size $K \times V$, we add Gaussian noise to each component of this matrix:
\begin{align}\label{eq:gaussian_LDA}
\tilde{\bar{\vs}}_{k}^v = {\bar{\vs}}_{k}^v + Y_k^v, \mbox{ where } Y_k^v \sim \Nrm(0, \sigma^2 (\Delta \bar{\vs})^2), 
\end{align}
where 
$\bar{\vs}_{k}^v = \tfrac{1}{S} \sum_d \sum_n \vphi_{dn}^k \vw_{dn}^v$, and $\Delta \bar{\vs}$ is the sensitivity.  We then map the perturbed components to 0 if they become negative.  For LDA, the worst-case sensitivity is given by
\begin{align}\label{eq:sen_lap_lda}
\Delta \bar{\vs} 
&= \max_{|\Dat \setminus {\Dat'}|=1} \sqrt{ \sum_k \sum_v (\bar{\vs}_{k}^v(\Dat)-\bar{\vs}_{k}^v(\Dat'))^2}, \nonumber \\
&= \max_{|\Dat \setminus {\Dat'}|=1}  \sqrt{\sum_k \sum_v  \left( \frac{1}{
S}\sum_n \sum_{d=1}^S \vphi_{dn}^k \vw_{dn}^v - \frac{1}{S-1} \sum_n \sum_{d=1}^{S-1}\vphi_{dn}^k \vw_{dn}^v \right)^2}, \nonumber \\
 &=  \max_{|\Dat \setminus {\Dat'}|=1}  \sqrt{\sum_k \sum_v \left|\frac{S-1}{S} \frac{1}{S-1} \sum_n \sum_{d=1}^{S-1} \vphi_{dn}^k \vw_{dn}^v + \frac{1}{S} \sum_n \vphi_{Sn}^k \vw_{Sn}^v - \frac{1}{S-1} \sum_n \sum_{d=1}^{S-1}\vphi_{dn}^k \vw_{dn}^v \right|^2}, \nonumber \\
  &=  \max_{\vphi_{Sn}^k, \vw_{Sn}^v}  \sqrt{\sum_k \sum_v \left|\frac{1}{S} \sum_n \vphi_{Sn}^k \vw_{Sn}^v -\frac{1}{S} \left( \frac{1}{S-1} \sum_n \sum_{d=1}^{S-1}\vphi_{dn}^k \vw_{dn}^v\right) \right|^2}, \nonumber \\
    &=  \max_{\vphi_{Sn}^k, \vw_{Sn}^v}  \sqrt{\sum_k \sum_v \left|\frac{1}{S} \sum_n \vphi_{Sn}^k \vw_{Sn}^v\right|^2}, \nonumber \\
    & \quad \mbox{ since $0 \leq \vphi_{dn}^k \leq 1$, $\vw_{dn}^v \in \{0, 1 \}$, and we assume $0 \leq \frac{1}{S-1} \sum_n \sum_{d=1}^{S-1}\vphi_{dn}^k \vw_{dn}^v \leq \sum_n \vphi_{Sn}^k \vw_{Sn}^v $,  }\nonumber \\
 &\leq \max_{\vphi_{Sn}^k, \vw_{Sn}^v} \; \; \frac{1}{S} \sum_n (\sum_k\vphi_{Sn}^k) (\sum_v \vw_{Sn}^v)  \leq \frac{N}{S}, \end{align}
since  $\sum_k\vphi_{Sn}^k = 1$, and $\sum_v \vw_{Sn}^v=1$.  This sensitivity accounts for the worst case in which all $N$ words in $\vw_{S}$ are assigned to the same entry of $\bar{\vs}$, i.e. they all have the same word type $v$, and are hard-assigned to the same topic $k$ in the variational distribution.  In our practical implementation, we improve the sensitivity by exploiting the fact that most typical documents' normalized expected sufficient statistic matrices $\bar{\vs}^d_{vk} = \frac{1}{S}\sum_n \vphi_{dn}^k \vw_{dn}^v$ (a $K \times V$ matrix for document $d$) have a much smaller norm than this worst case.  Specifically, inspired by \cite{2016arXiv160700133A}, we apply a norm clipping strategy, in which the contribution $\bar{\vs}^d$ of each document to the sufficient statistics matrix $\bar{\vs} = \sum_d\bar{\vs}^d$ is clipped (or projected) such that the Frobenious norm of the matrix is bounded by $|\bar{\vs}^d| \leq a \frac{N}{S}$, for a user-specified $a \in (0,1]$.  For each document, if this criterion is not satisfied, we project the expected sufficient statistics down to the required norm via
\begin{equation}
\bar{\vs}^d := a \frac{N}{S}\frac{\bar{\vs}^d}{{|\bar{\vs}^d|}} \mbox{ .} 
\end{equation}
After this the procedure, the sensitivity of the entire matrix becomes $a \Delta \bar{\vs}$ (i.e., $a N/S$), and we add noise on this scale to the clipped expected sufficient statistics.  We set $a = 0.1$ in our experiments, which empirically resulted in clipping being applied to around $\frac{1}{2}$ to $\frac{3}{4}$ of the documents.
The resulting algorithm is summarised in \algoref{PPVI_LDA_minibatches}.

\begin{algorithm}[t]
\caption{VIPS for LDA}
\label{algo:PPVI_LDA_minibatches}
\begin{algorithmic}
\REQUIRE Data $\Dat$. Define $D$ (documents), $V$ (vocabulary), $K$ (number of topics).\\
\quad \quad \;   Define $\rho_{t} = (\tau_0 + t)^{-\kappa}$, mini-batch size $S$, hyperparameters $\alpha, \eta$, $\sigma^2$, and $a$
\ENSURE Privatised expected natural parameters $\langle \log \vbeta_k \rangle_{q(\vbeta_k)}$ and sufficient statistics $\tilde{\bar{\vs}}$.
\STATE Compute the sensitivity of the expected sufficient statistics given in \eqref{sen_lap_lda}.
\FOR{$t = 1, \ldots, J $}
\STATE {\it{\textbf{(1) E-step}}}: Given expected natural parameters $\langle \log \vbeta_k \rangle_{q(\vbeta_k)}$
\STATE $\bar{\vs} := \mathbf{0}$
\FOR{$d=1,\ldots,S$}
   \FOR{$r=1,\ldots,R$}
	   \STATE Compute $q(\vz_{dn}^k)$ parameterised by $\vphi_{dn}^k \propto \exp(\langle \log \vbeta_k \rangle_{q(\vbeta_k)} \trp \vw_{dn} + \langle \log \vtheta_d^k \rangle_{q(\vtheta_d)})$.
	   \STATE Compute $q(\vtheta_d)$ parameterised by $\vgamma_d = \alpha \vone_K + \sum_{n=1}^N \langle \vz_{dn}\rangle_{q(\vz_{dn})}$.
   \ENDFOR
   \STATE $\bar{\vs}^d_{vk} := \frac{1}{S}\sum_n \vphi_{dn}^k \vw_{dn}^v$
   \IF{$|\bar{\vs}^d| > aN/S$}
      \STATE $\bar{\vs}^d := a \frac{ N}{S}\frac{\bar{\vs}^d}{{|\bar{\vs}^d|}}$ 
   \ENDIF
   \STATE $\bar{\vs} := \bar{\vs} + \bar{\vs}^d$
\ENDFOR
%
\STATE Output the perturbed expected sufficient statistics $ \tilde{\bar{\vs}}_k^v = \bar{\vs} + Y_k^v$, where $Y_k^v$ is Gaussian noise given in \eqref{gaussian_LDA}, but using sensitivity $aN/S$.
\STATE Clip negative entries of $\tilde{\bar{\vs}}$ to 0.
\STATE Update the log-moment functions.
\STATE {\it{\textbf{(2) M-step}}}: Given perturbed expected sufficient statistics $ \tilde{\bar{\vs}}_k $,
\STATE Compute $q(\vbeta_k)$ parameterised by $\vlambda_k^{(t)} = \eta \vone_V + D \tilde{\bar{\vs}}_k$.
\STATE Set $\vlambda^{(t)} \mapsfrom (1-\rho_t)\vlambda^{(t-1)} + \rho_t \vlambda^{(t)} $.
\STATE Output expected natural parameters $\langle \log \vbeta_k \rangle_{q(\vbeta_k)}$.
\ENDFOR
\end{algorithmic}
\end{algorithm}

\begin{figure}[t]
	\centering
	\centerline{\includegraphics[width=1\textwidth]{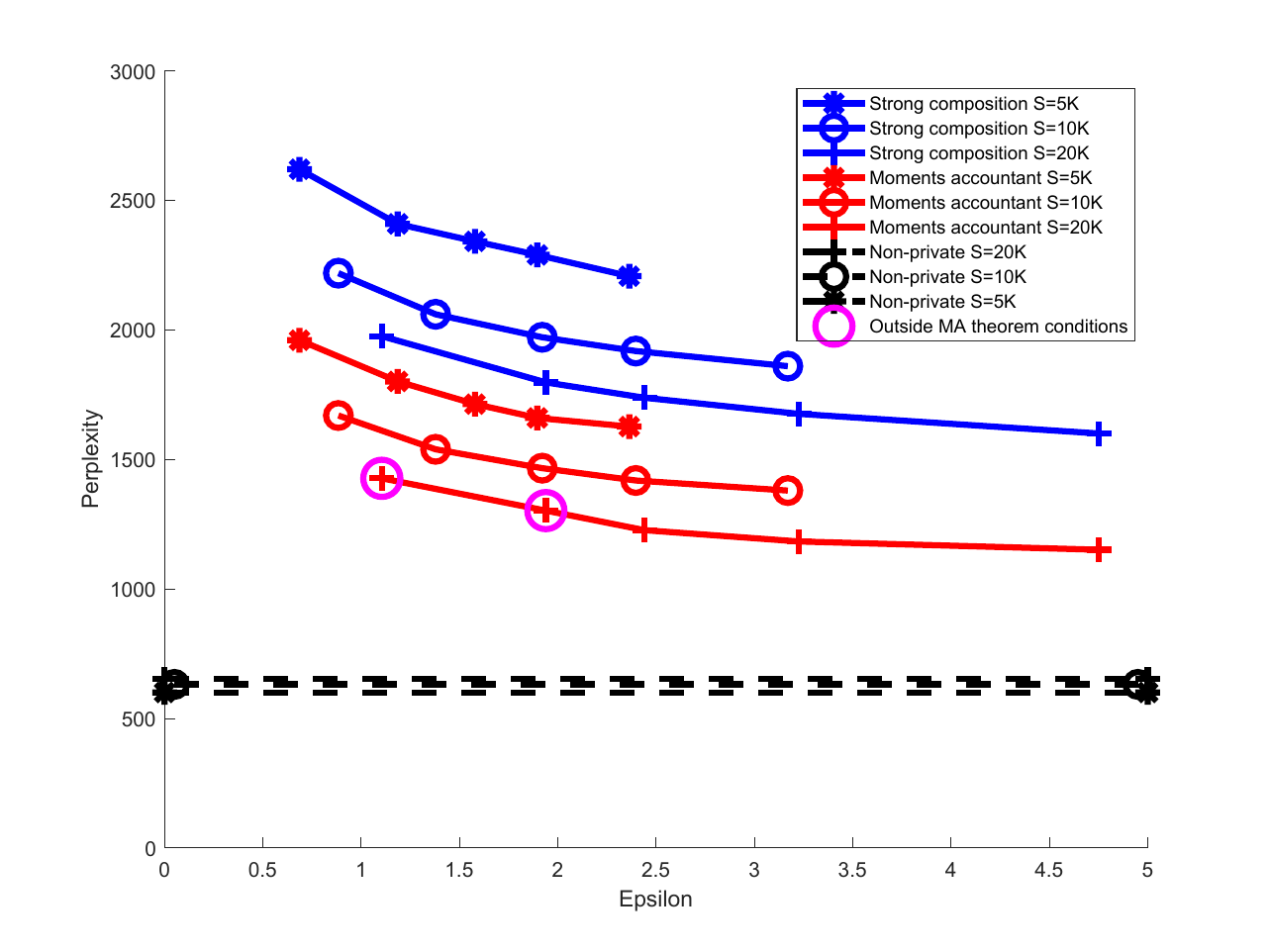}}
	\caption{ Epsilon versus perplexity, varying $\sigma$ and $S$, Wikipedia data, one epoch.  The parameters for the two data points indicated by the pink circles do not satisfy the conditions of the moments accountant composition theorem, so those $\epsilon$ values are not formally proved.	}
	\label{fig:epsilonVsPerplexity}
\end{figure}

\begin{figure}[t]
	\centering
	\centerline{\includegraphics[width=0.7\textwidth]{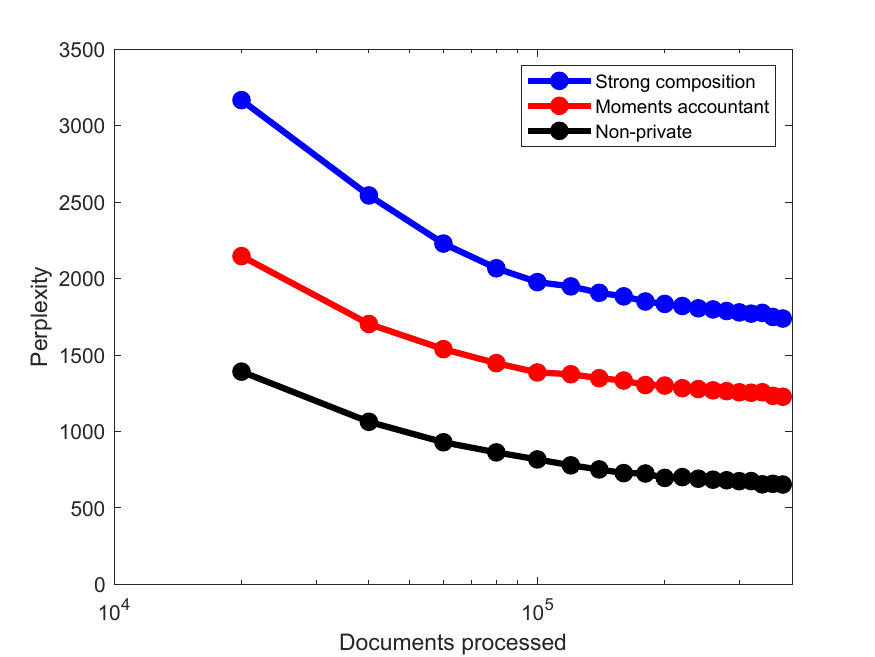}}
	\caption{ Perplexity per iteration. Wikipedia data, $S=20,000$, $\epsilon=2.44$, one epoch.	}
	\label{fig:PerplexityVsIter}
\end{figure}

\subsection{Experiments using Wikipedia data}

We downloaded a random $D=400,000$ documents from Wikipedia to test our VIPS algorithm.  We used $50$ topics and a vocabulary set of approximately $8000$ terms.  The algorithm was run for one epoch in each experiment.

We compared our moments accountant approach with a baseline method using the {\it{strong composition}}
%
(Theorem 3.20 of \citet{Dwork14}), resulting from the max divergence of the privacy loss random variable being bounded by a total budget including a slack variable $\delta$, which yields
$(J\epsilon' (e^{\epsilon'}-1) + \sqrt{2J\log(1/\delta{''})}\epsilon', \; \delta{''} + J\delta')$-DP.

As our evaluation metric, we compute an upper bound on the perplexity on held-out documents.  Perplexity is an information-theoretic measure of the predictive performance of probabilistic models which is commonly used in the context of language modeling \citep{jelinek1977perplexity}.  The perplexity of a probabilistic model $p_{model}(x)$ on a test set of $N$ data points $x_i$ (e.g. words in a corpus) is defined as
\begin{equation}
b^{-\frac{1}{N}\sum_{i=1}^N \log_b p_{model}(x_i)} \mbox{ ,} \label{eqn:perplexity}
\end{equation}
where $b$ is generally either $2$ or $e$, corresponding to a measurement based on either bits or nats, respectively.  We can interpret $-\frac{1}{N}\sum_{i=1}^N \log_b p_{model}(x_i)$ as the cross-entropy between the model and the empirical distribution.  This is the expected number of bits (nats) needed to encode a data point from the empirical data distribution (i.e. a word in our case), under an optimal code based on the model.  Perplexity is $b$ to the power of the cross entropy, which converts the number of bits (nats) in the encoding to the number of possible values in an encoding of that length (supposing the cross entropy were integer valued).  Thus, perplexity measures the \emph{effective vocabulary size when using the model to encode the data}, which is understood as a reflection of how confused (i.e. ``perplexed'') the model is.  In our case, $p_{model}$ is the posterior predictive distribution under our variational approximation to the posterior, which is intractable to compute. Following \citet{NIPS2010_3902}, we approximate perplexity based on the learned variational distribution, measured in nats, by plugging the ELBO into Equation \ref{eqn:perplexity}, which results in an upper bound:
\begin{equation}
\mbox{perplexity}(\Dat^{test}, \vlambda) \leq \exp \left[ - \left(\sum_i \langle \log p(\vn^{test}, \vtheta_i, \vz_{i}|\vlambda) \rangle_{q(\vtheta_i, \vz_i)} -  \langle \log q(\vtheta, \vz) \rangle_{q(\vtheta, \vz)} \right) / \sum_{i,n} \vn_{i,n}^{test} \right], \nonumber 
\end{equation}
 where $\vn_i^{test}$ is a vector of word counts for the $i$th document, $\vn^{test} = \{ \vn_i^{test} \}_{i=1}^I$. In the above, we use the $\vlambda$ that was calculated during training. We compute the posteriors over $\vz$ and $\vtheta$ by performing the first step in our algorithm using the test data and the perturbed sufficient statistics we obtain during training. We adapted the python implementation by the authors of \citep{NIPS2010_3902} for our experiments.
 
Figure \ref{fig:epsilonVsPerplexity} shows the trade-off between $\epsilon$ and per-word perplexity on the Wikipedia dataset for the different methods under a variety of conditions, in which we varied the value of $\sigma \in \{1.0, 1.1,1.24,1.5,2\}$ and the minibatch size $S \in \{5,000, 10,000, 20,000\}$.  We found that the moments accountant composition substantially outperformed strong composition in each of these settings.  Here, we used relatively large minibatches, which were necessary to control the signal-to-noise ratio in order to obtain reasonable results for private LDA. Larger minibatches thus had lower perplexity.  However, due to its impact on the subsampling rate, increasing $S$ comes at the cost of a higher $\epsilon$ for a fixed number of documents processed (in our case, one epoch).  The minibatch size $S$ is limited by the conditions of the moments accountant composition theorem shown by \cite{2016arXiv160700133A}, with the largest valid value being obtained at around $S\approx20,000$ for the small noise regime where $\sigma \approx 1$.  To show the convergence behavior of the algorithms, we also report the perplexity per iteration in Figure \ref{fig:PerplexityVsIter}.  We found that the gaps between methods remain relatively constant at each stage of the learning process.  Similar results were observed for other values of $\sigma$ and $S$.

In Table \ref{my-table}, for each method, we show the top $10$ words in terms of assigned probabilities for $3$ example topics. 
Non-private LDA results in the most coherent words among all the methods. 
For the private LDA models with a total privacy budget $\epsilon=2.44$ ($S=20,000, \sigma=1.24$), as we move from moments accountant to strong composition, the amount of noise added gets larger, and the topics become less coherent.  We also observe that the probability mass assigned to the most probable words decreases with the noise, and thus strong composition gave less probability to the top words compared to the other methods.

\begin{table}[t]
	\centering
	\caption{Posterior topics from private ($\epsilon=2.44$) and non-private LDA}
	\label{my-table}
	\begin{tabular}{llllll}
		\toprule
		Non-private &        & Moments Acc. &        & Strong Comp. &        \\ 
		\cmidrule(r){1-2} \cmidrule(l){3-4} \cmidrule(l){5-6}
		topic 33:   &        & topic 33:   &        & topic 33:   &        \\
		german      & 0.0244 & function    & 0.0019 & resolution  & 0.0003 \\
		system      & 0.0160 & domain      & 0.0017 & northward   & 0.0003 \\
		group       & 0.0109 & german      & 0.0011 & deeply      & 0.0003 \\
		based       & 0.0089 & windows     & 0.0011 & messages    & 0.0003 \\
		science     & 0.0077 & software    & 0.0010 & research    & 0.0003 \\
		systems     & 0.0076 & band        & 0.0007 & dark        & 0.0003 \\
		computer    & 0.0072 & mir         & 0.0006 & river       & 0.0003 \\
		software    & 0.0071 & product     & 0.0006 & superstition& 0.0003 \\
		space       & 0.0061 & resolution  & 0.0006 & don         & 0.0003 \\
		power       & 0.0060 & identity    & 0.0005 & found       & 0.0003 \\
		&        &             &        &             &        \\
		topic 35:   &        & topic 35:   &        & topic 35:   &        \\
		station     & 0.0846 & station     & 0.0318 & station     & 0.0118 \\
		line        & 0.0508 & line        & 0.0195 & line        & 0.0063 \\
		railway     & 0.0393 & railway     & 0.0149 & railway     & 0.0055 \\
		opened      & 0.0230 & opened      & 0.0074 & opened      & 0.0022 \\
		services    & 0.0187 & services    & 0.0064 & services    & 0.0015 \\
		located     & 0.0163 & closed      & 0.0056 & stations    & 0.0015 \\
		closed      & 0.0159 & code        & 0.0054 & closed      & 0.0014 \\
		owned       & 0.0158 & country     & 0.0052 & section     & 0.0013 \\
		stations    & 0.0122 & located     & 0.0051 & platform    & 0.0012 \\
		platform    & 0.0109 & stations    & 0.0051 & company     & 0.0010 \\
		&        &             &        &             &        \\
		topic 37:   &        & topic 37:   &        & topic 37:   &        \\
		born        & 0.1976 & born        & 0.0139 & born        & 0.0007 \\
		american    & 0.0650 & people      & 0.0096 & american    & 0.0006 \\
		people      & 0.0572 & notable     & 0.0092 & street      & 0.0006 \\
		summer      & 0.0484 & american    & 0.0075 & charles     & 0.0004 \\
		notable     & 0.0447 & name        & 0.0031 & said        & 0.0004 \\
		canadian    & 0.0200 & mountain    & 0.0026 & events      & 0.0004 \\
		event       & 0.0170 & japanese    & 0.0025 & people      & 0.0003 \\
		writer      & 0.0141 & fort        & 0.0025 & station     & 0.0003 \\
		dutch       & 0.0131 & character   & 0.0019 & written     & 0.0003 \\
		actor       & 0.0121 & actor       & 0.0014 & point       & 0.0003 \\
		\bottomrule
	\end{tabular}
\end{table}



%

\section{VIPS for non-CE family }\label{non-CE-family}

Under non-CE family models, the complete-data likelihood typically has the following form: 
\begin{equation}
p(\Dat_n, \vl_n| \vm) \propto \exp(-h(\vm, \vs(\Dat_n, \vl_n))) \mbox{ ,  }
\end{equation} which includes some function $h(\vm, \vs(\Dat_n, \vl_n))$ that cannot be split into two functions, where one is a function of only $\vm$ and the other is a function of only $\vs(\Dat_n, \vl_n)$. Hence, we cannot apply the general VIPS framework we described in the previous section to this case. 

However, when the models we consider have binomial likelihoods, for instance, under negative binomial regression, nonlinear mixed-effects models, spatial models for count data, and logistic regression, we can bring the non-CE models to the CE family by adopting the P{\'o}lya-Gamma data augmentation strategy introduced by \citet{PolsonScott13}.
%
%
\paragraph{P{\'o}lya-Gamma data augmentation}
%
P{\'o}lya-Gamma data augmentation introduces an auxiliary variable that is   P{\'o}lya-Gamma distributed per datapoint, such that the log-odds can be written as mixtures of Gaussians with respect to a P{\'o}lya-Gamma distribution, as stated in Theorem 1 in \citep{PolsonScott13}:
\begin{eqnarray}
\frac{\exp(\psi_n)^{y_n}}{(1+\exp(\psi_n))^b} 
&=& 2^{-b} \exp((y_n-\tfrac{b}{2})\psi_n) \int_{0}^\infty \exp(-\tfrac{\xi_n\psi_n^2}{2}) \; p(\xi_n) \; d\xi_n
\end{eqnarray} where $\psi_n$ is a linear function in model parameters $\vm$,  $y_n$ is the $n$th observation, and $\xi_n$ is a P{\'o}lya-Gamma random variable, $ \xi_n \sim \mbox{PG}(b, 0)$ where $b>0$. For example, $\psi_n=\vm\trp\vx_n$ for models without latent variables and $\vx_n$ is the $n$th input vector, or $\psi_n=\vm\trp \vl_n$ for models with latent variables in unsupervised learning. Note that $b$ is set depending on which binomial model one uses. For example, $b=1$ in logistic regression. When $\psi_n=\vl_n \trp \vm$ and $b=1$, we  
express the likelihood as: 
\begin{align}\label{eq:PG_AUG}
p(\Dat_n|\vl_n, \vm) &= 2^{-1}\exp((y_n-\tfrac{1}{2})\vl_n\trp \vm) \int_0^\infty \exp(-\tfrac{1}{2}\xi_n\vl_n\trp \vm \vm\trp \vl_n) p(\xi_n) d\xi_n.
\end{align}  
By introducing a variational posterior over the auxiliary variables, we introduce a new objective function (See Appendix A for derivation)
\begin{align}\label{eq:lower_lower_bound_PG}
&\mathcal{L}_n (q(\vm), q(\vl_n), q(\xi_n)) \nonumber \\
& = \int \; q(\vm) q(\vl_n) q(\xi_n) \log \frac{p(\Dat_n|\vl_n, \xi_n, \vm )p(\vl_n)p(\xi_n)p(\vm)}{q(\vm)q(\vl_n) q(\xi_n)}, \\
&= -\log 2 + (y_n - \tfrac{1}{2}) \langle \vl_n \rangle_{q(\vl_n)} \trp \langle \vm \rangle_{q(\vm)} -\tfrac{1}{2} \langle \xi_n \rangle_{q(\xi_n)} \langle \vl_n\trp \vm \vm\trp \vl_n \rangle_{q(\vl_n)q(\vm)} , \nonumber \\
& \qquad -  \mbox{D}_{KL}(q(\xi_n)||p(\xi_n)) - \mbox{D}_{KL}(q(\vm)||p(\vm)) - \mbox{D}_{KL}(q(\vl_n)||p(\vl_n)). \nonumber 
\end{align} 
The first derivative of the {{lower}} lower bound with respect to $q(\xi_n)$ gives us a closed form update rule
\begin{align}
\frac{\partial}{\partial q(\xi_n)}  \mathcal{L}_n (q(\vm), q(\vl_n), q(\xi_n)) =0, \; 
&\; \mapsto q(\xi_n) \propto \mbox{PG}(1, \sqrt{ \langle \vl_n\trp \vm \vm\trp \vl_n \rangle_{q(\vl_n)q(\vm)} } ).
\end{align} The rest updates for $q(\vl_n) q(\vm)$ follow the standard updates under the conjugate exponential family distributions, since the lower bound to the conditional likelihood term includes only linear and quadratic terms both in $\vl_n$ and $\vm$.
By introducing the auxiliary variables, the complete-data likelihood conditioned on $\xi_n$ (with some prior on $\vl_n$) now has the form of 
 \begin{align}\label{eq:aux_complete_data_like}
p(\Dat_n, \vl_n| \vm, \xi_n) &\propto p(\Dat_n| \vl_n, \vm, \xi_n) \; p(\vl_n), \nonumber \\
&\propto \exp(\vn(\vm)\trp \vs(\Dat_n, \vl_n, \xi_n)) \; p(\vl_n),
\end{align} which consists of natural parameters and sufficient statistics given by 
\begin{equation}
 \vn(\vm) = \begin{bmatrix}
       \vm     \\[0.3em]
       \mbox{vec}(- \frac{1}{2}\vm \vm\trp)          \\[0.3em]
     \end{bmatrix}, 
     \quad 
     \vs(\Dat_n, \vl_n, \xi_n) =  \begin{bmatrix}
        (y_n-\tfrac{1}{2}) \vl_n    \\[0.3em]
       \mbox{vec}(\xi_n \vl_n \vl_n \trp)          \\[0.3em]
     \end{bmatrix}.
\end{equation} Note that now not only the latent and observed variables but also the new variables $\xi_i$ form the complete-data sufficient statistics. 
%
%
The resulting variational Bayes algorithm for models with binomial likelihoods is given by 
\begin{align}
&(a)\mbox{ Given  the expected natural parameters $\bar{\vn}$, the E-step yields: }    \nonumber \\
& \qquad  q(\vl, \vxi)  = \prod_{n=1}^N q(\vl_n)q(\xi_n) \propto p(\vxi) \exp \left[ \int d\vm  \; q(\vm) \log p(\Dat, \vl|\vm, \vxi) \right], \nonumber \\
& \qquad \qquad \qquad \qquad \qquad \qquad \propto p(\vxi) p(\vl) \prod_{n=1}^N \exp(\bar{\vn}\trp  \vs(\Dat_n, \vl_n, \xi_n) ) \nonumber,  \\
& \qquad \mbox{where } q(\xi_n) = \mbox{PG}(1, \sqrt{\langle\vl_i\trp \vm \vm\trp \vl_n \rangle_{q(\vl_n, \vm)}}), \mbox{ and } p(\xi_n) = \mbox{PG}(1, 0) \\
& \qquad \qquad \; \; \; q(\vl_n) \propto  p(\vl_n) \exp(\bar{\vn}\trp  \langle \vs(\Dat_n, \vl_n, \xi_n)\rangle_{q(\xi_n)} ). \\
& \qquad \mbox{Using $q(\vl)q(\xi)$, it outputs $\bar{\vs}(\Dat) = \tfrac{1}{N}\sum_{n=1}^N \langle {\vs}(\Dat_n, \vl_n, \xi_n) \rangle_{q(\vl_n) q(\xi_n)} $.} \nonumber \\
&(b)\mbox{ Given the expected sufficient statistics $\bar{\vs}$,  the M-step yields: }  \nonumber \\
&  \qquad q(\vm) \propto p(\vm) \exp\left[\int d\vl \;  d\vxi \; q(\vl) q(\vxi)\; \log p(\Dat, \vl|\vm, \vxi) \right] , \nonumber \\
& \qquad \qquad \propto \exp(\vn(\vm)\trp \tilde{\vnu}), \mbox{ where } \; \tilde{\vnu} = \vnu + N \bar{\vs}(\Dat).\\
& \qquad \mbox{Using $q(\vm)$, it outputs the expected natural parameters $\bar{\vn}:= \langle \vn(\vm) \rangle_{q(\vm)}$.}\nonumber 
\end{align} 
Similar to the VIPS algorithm for the CE family, perturbing the expected sufficient statistics  $\bar{\vs}(\Dat)$ in the E-step suffices for privatising all the outputs of the algorithm. 
Algorithm \ref{algo:VIPS_for_nonCE} summarizes private stochastic variational Bayes algorithm for non-CE family with binomial likelihoods. 
\begin{algorithm}[h]
\caption{($\epsilon_{tot}, \delta_{tot}$)-DP VIPS for non-CE family with binomial likelihoods}
\label{algo:VIPS_for_nonCE}
\begin{algorithmic}
\REQUIRE Data $\Dat$. Define $\rho_{t} = (\tau_0 + t)^{-\kappa}$,  mini-batch size $S$, maximum iterations $J$, and $\sigma$.
\ENSURE Perturb expected natural parameters $\tilde{\bar{\vn}}$ and expected sufficient stats $\tilde{\bar{\vs}}$.
\STATE Compute the L2-sensitivity $\Delta$ of the expected sufficient statistics.
\FOR{$t = 1, \ldots, J $}
\STATE {\it{\textbf{(1) E-step}}}: Given expected natural parameters $\bar{\vn}$, compute $q(\vl_{n})q(\xi_n)$
for\\ $n=1,\ldots,S$. Perturb each coordinate of $\bar{\vs} = \tfrac{1}{S} \sum_{n=1}^S \langle \vs( \Dat_n, \vl_n) \rangle_{q(\vl_{n})q(\xi_n)}$ by adding noise drawn from $\Nrm(0, \sigma^2 \Delta^2)$. Output $ \tilde{\bar{\vs}} $.
\STATE Update the moments function.
\STATE {\it{\textbf{(2) M-step}}}: Given $ \tilde{\bar{\vs}} $,
compute $q(\vm)$ by $\tilde{\vnu}^{(t)} = \vnu + N \tilde{\bar{\vs}} $.
Set $\tilde{\vnu}^{(t)} \mapsfrom (1-\rho_t)\tilde{\vnu}^{(t-1)} + \rho_t \tilde{\vnu}^{(t)} $. 
Output the expected natural parameters $\tilde{\bar{\vn}} = \langle \vm \rangle_{q(\vm)}$.
\ENDFOR
\end{algorithmic}
\end{algorithm}

As a side note, in order to use the variational lower bound as a stopping criterion, one needs to draw samples from the P{\'o}lya-Gamma posterior to numerically calculate the lower bound given in \eqref{lower_lower_bound_PG}.  This might be a problem if one does not have access to the P{\'o}lya-Gamma posterior, since our algorithm only outputs the perturbed expected sufficient statistics in the E-step (not the P{\'o}lya-Gamma posterior itself).
However, one could use other stopping criteria, which do not require sampling from the P{\'o}lya-Gamma posterior, e.g., calculating the prediction accuracy in the classification case.

\section{VIPS for Bayesian logistic regression}\label{bayesian_logistic_reg}


We present an example of non-CE family, Bayesian logistic regression, and illustrate how to employ the VIPS framework given in Algorithm \ref{algo:VIPS_for_nonCE} in such a case.  


\subsection{Model specifics}
Under the logistic regression model with the Gaussian prior on the weights $\vm \in\mathbb{R}^d$,
\begin{eqnarray}
p(y_n=1|\vx_n, \vm) = \sigma(\vm\trp\vx_n), \quad p(\vm|\alpha) = \Nrm(\vm|0, \alpha^{-1} I), \quad  
p(\alpha) = \mbox{Gam}(a_0, b_0),
\end{eqnarray} 
where $\sigma(\vm\trp\vx_n) = 1/({1+ \exp(-\vm\trp\vx_n)}) $, the $n$th input is $\vx_n\in\mathbb{R}^d$, and the $n$th output is $y_n \in \{0, 1 \}$.
In logistic regression, there are no latent variables. Hence, the complete-data likelihood coincides the data likelihood,
\begin{eqnarray}
p(y_n|\vx_n, \vm) &=& \frac{\exp(\psi_n)^{y_n}}{1+\exp(\psi_n)}, 
\end{eqnarray} where $\psi_n = \vx_n\trp \vm$. Since the likelihood is not in the CE family. 
We use the P{\'o}lya-Gamma data augmentation trick to re-write it as
\begin{eqnarray}
p(y_n|\vx_n, \xi_n, \vm) &\propto& \exp((y_n-\tfrac{1}{2})\vx_n\trp \vm)\exp(-\tfrac{1}{2}\xi_n\vx_n\trp \vm \vm\trp \vx_n), 
\end{eqnarray}  
and the data likelihood conditioned on $\xi_n$ is 
 \begin{equation}\label{eq:aux_complete_data_like_logistic_reg} 
p(y_n|\vx_n, \vm, \xi_n) \propto \exp(\vn(\vm)\trp \vs(\Dat_n, \xi_n)),
\end{equation} where the natural parameters and sufficient statistics are given by 
\begin{equation}
 \vn(\vm) = \begin{bmatrix}
       \vm     \\[0.3em]
       \mbox{vec}(- \frac{1}{2}\vm \vm\trp) \\[0.3em]
     \end{bmatrix}, 
     \quad 
     \vs(\Dat_n, \xi_n) =  \begin{bmatrix}
        (y_n-\tfrac{1}{2}) \vx_n    \\[0.3em]
       \mbox{vec}(\xi_n \vx_n \vx_n \trp)           \\[0.3em]
     \end{bmatrix}.
\end{equation} 

\paragraph{Variational Bayes in Bayesian logistic regression}

Using the likelihood given in \eqref{aux_complete_data_like}, we compute the posterior distribution over $\vm, \alpha, \vxi$ by maximising the following variational lower bound  due to \eqref{lower_lower_bound_PG}
\begin{align}\label{eq:aux_lbd_final_logistic}
&\mathcal{L} (q(\vm), q(\vxi), q(\alpha)) = \sum_{n=1}^N \left[ - \log 2 + (y_n - \tfrac{1}{2}) \vx_n  \trp \langle \vm \rangle_{q(\vm)} -\tfrac{1}{2} \langle \xi_n \rangle_{q(\xi_n)}\vx_n\trp  \langle \vm \vm\trp\rangle_{q(\vm)} \vx_n  \right], \nonumber \\
& \quad \qquad \qquad \qquad \qquad - \sum_{n=1}^N \mbox{D}_{KL}(q(\xi_n)||p(\xi_n)) - \mbox{D}_{KL}(q(\vm)||p(\vm)) - \mbox{D}_{KL}(q(\alpha)||p(\alpha)). \nonumber 
\end{align} 
%
%
In the E-step, we update
\begin{align}
q(\vxi) &\propto p(\vxi)\prod_{n=1}^N \exp(\bar{\vn} (\vm)\trp \vs(\Dat_n, \xi_n) ) , \\
&= \prod_{n=1}^N q(\xi_n), \mbox{ where } q(\xi_n) = \mbox{PG}(1, \sqrt{\vx_n\trp \langle  \vm \vm\trp \rangle_{q(\vm)}\vx_n}).
\end{align} 
Using $q(\vxi)$, we compute the expected sufficient statistics 
\begin{eqnarray}
\bar{\vs}= \tfrac{1}{N} \sum_{n=1}^N \bar{\vs}(\Dat_n) , \mbox{ where }
\bar{\vs}(\Dat_n) =  
\begin{bmatrix}
        \bar{\vs}_1(\Dat_n)    \\[0.3em]
        \bar{\vs}_2(\Dat_n)        \\[0.3em]
     \end{bmatrix}
= \begin{bmatrix}
        \tfrac{1}{N} (y_n-\tfrac{1}{2}) \vx_n    \\[0.3em]
        \tfrac{1}{N} \mbox{vec} (\langle \xi_n  \rangle_{q(\xi_n)} \vx_n \vx_n \trp )          \\[0.3em]
     \end{bmatrix}.
\end{eqnarray}

\begin{algorithm}[t]
\caption{VIPS for Bayesian logistic regression}
\label{algo:VIPS_BLR}
\begin{algorithmic}
\REQUIRE Data $\Dat$. Define $\rho_{t} = (\tau_0 + t)^{-\kappa}$, mini-batch size $S$, and maximum iterations $J$
\ENSURE Privatised expected natural parameters $\tilde{\bar{\vn}}$ and expected sufficient statistics $\tilde{\bar{\vs}}$
\STATE Using the sensitivity of the expected sufficient statistics given in \eqref{Mean_sens} and \eqref{Cov_sens}, 
\FOR{$t = 1, \ldots, J $}
\STATE {\it{\textbf{(1) E-step}}}: Given expected natural parameters $\bar{\vn}$, compute $q(\xi_n)$
for $n=1,\ldots,S$. \\
Perturb $\bar{\vs} = \tfrac{1}{S} \sum_{n=1}^S \langle \vs( \Dat_n, \xi_n) \rangle_{q(\xi_n)}$ by \eqref{perturbmean_Gaussian} and \eqref{perturbcov}, and output $ \tilde{\bar{\vs}} $.
\STATE Update the moments function.
\STATE {\it{\textbf{(2) M-step}}}: Given $ \tilde{\bar{\vs}} $,
compute $q(\vm)$ by $\tilde{\vnu}^{(t)} = \vnu + N \tilde{\bar{\vs}} $.
Set $\tilde{\vnu}^{(t)} \mapsfrom (1-\rho_t)\tilde{\vnu}^{(t-1)} + \rho_t \tilde{\vnu}^{(t)} $.
Using $\tilde{\vnu}^{(t)}$, update $q(\alpha)$ by \eqref{alpha_update}, and output $\tilde{\bar{\vn}} = \langle \vm \rangle_{q(\vm)}$.
\ENDFOR
\end{algorithmic}
\end{algorithm}

In the M-step, we compute $q(\vm)$ and $q(\alpha)$ by 
\begin{align}
q(\vm) &\propto p(\Dat|\vm, \langle \vxi \rangle) \;  p(\vm|\langle \alpha\rangle) \propto \exp(\vn(\vm)\trp \tilde{\vnu} ) = \Nrm(\vm| \vmu_{\vm}, \Sigma_{\vm}),  \\
q(\alpha) &\propto  p(\vm|\alpha) p(\alpha|a_0, b_0) = \mbox{Gamma}(a_N, b_N), 
\end{align}
where $\tilde{\vnu} = \vnu + N \bar{\vs}$, $\vnu = [\mathbf{0}_d, \langle \alpha\rangle_{q(\alpha)} I_d ]$, and 
\begin{align}\label{eq:alpha_update}
a_N = a_0 + \tfrac{d}{2}, \quad b_N = b_0 + \tfrac{1}{2}(\vmu_{\vm}\trp\vmu_{\vm} + \mbox{tr}(\Sigma_{\vm}) ).
\end{align}
Mapping from $\tilde{\vnu}$ to $(\vmu_{\vm}, \Sigma_{\vm})$ is deterministic, as below, where $\bar{\vs}_1 = \sum_{n=1}^N \bar{\vs}_1(\Dat_n)$ and $\bar{\vs}_2 = \sum_{n=1}^N \bar{\vs}_2(\Dat_n)$, 
\begin{eqnarray} 
\tilde{\vnu}  = 
\begin{bmatrix}
        N {\bar{\vs}}_1 +  \mathbf{0}_d \\[0.3em]
       N {\bar{\vs}}_2 + \langle \alpha\rangle_{q(\alpha)} I_d         \\[0.3em]
     \end{bmatrix}
     = \begin{bmatrix}
       \Sigma_\vm^{-1} \vmu_\vm  \\[0.3em]
        \Sigma_\vm^{-1}      \\[0.3em]
     \end{bmatrix}.
\end{eqnarray}
Using $q(\vm)$, we compute expected natural parameters 
\begin{eqnarray} 
 \langle \vn(\vm) \rangle_{q(\vm)} 
 = \begin{bmatrix}
       \langle \vm    \rangle_{q(\vm)}  \\[0.3em]
       \mbox{vec}(-\frac{1}{2} \langle \vm \vm\trp \rangle_{q(\vm)}) \\[0.3em]
     \end{bmatrix}
     = \begin{bmatrix}
       \vmu_\vm  \\[0.3em]
        \mbox{vec}(-\frac{1}{2}(\Sigma_\vm +  \vmu_\vm \vmu_\vm\trp))      \\[0.3em]
     \end{bmatrix}.
\end{eqnarray}

\subsection{VIPS for Bayesian logistic regression}

Following the general framework of VIPS, we perturb the expected sufficient statistics. 
For perturbing $\bar{\vs}_1$, we add Gaussian noise to each coordinate, 
\begin{equation}\label{eq:perturbmean_Gaussian}
\tilde{\bar{\vs}}_1 = {\bar{\vs}}_1 + Y_{1, \ldots, d}, \mbox{ where  $Y_i \sim^{i.i.d.} \Nrm\left(0, \sigma^2 \Delta \bar{\vs}_1^2 \right)$}
\end{equation}
where the sensitivity $\Delta \bar{\vs}_1$ is given by
\begin{align}\label{eq:Mean_sens}
\Delta \bar{\vs}_1 &= \max_{|\Dat \setminus {\Dat'}|=1} |\bar{\vs}_1(\Dat) - \bar{\vs}_1({\Dat}')|_2 
\leq 
\max_{\vx_n, y_n} \tfrac{2}{N}|(y_n- \tfrac{1}{2})| \; |\vx_{n}|_2 \leq\tfrac{2}{N},
\end{align} due to \eqref{gen_sen} and the assumption that  the dataset is preprocessed such that any input has a maximum L2-norm of 1.

\begin{figure}[t]
\centering
\includegraphics[width=0.4\textwidth]{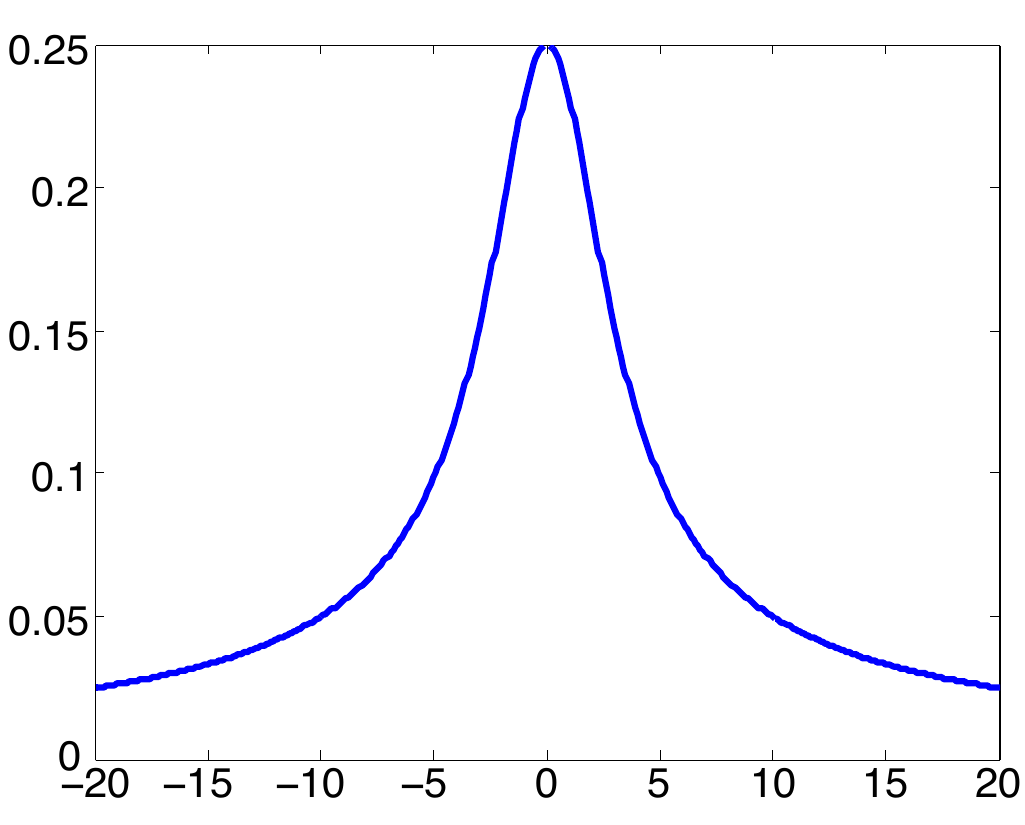}
  \caption{Posterior mean of each PG variable $\langle \xi_i \rangle$ as a function of $\sqrt{\vx_i\trp \langle  \vm \vm\trp \rangle\vx_i}$. 
The maximum value of $\langle \xi_i \rangle$ is 0.25 when $\sqrt{\vx_n\trp \langle  \vm \vm\trp \rangle\vx_n}= 0$,}
\label{fig:lambda}
\end{figure}

For  perturbing $\bar{\vs}_2$, we follow the \emph{Analyze Gauss} (AG) algorithm \citep{DworkTT014}. 
We first draw Gaussian random variables 
$\vz \sim \Nrm\left(0, \sigma^2 (\Delta\bar{\vs}_2)^2 I \right). $
Using $\vz$, we construct a upper triangular matrix (including diagonal), then copy the upper part to the lower part so that the resulting matrix $Z$ becomes
symmetric. Then, we add this noisy matrix to the covariance matrix 
\begin{equation}\label{eq:perturbcov}
\tilde{\bar{\vs}}_2 =  \bar{\vs}_2 + Z.
\end{equation} The perturbed covariance might not be positive definite. In such case, we project the negative eigenvalues to some value near zero to maintain positive definiteness of the covariance matrix. 
The sensitivity of $\bar{\vs}_2$ 
 in Frobenius norm is given by 
\begin{align}\label{eq:Cov_sens}
\Delta \bar{\vs}_2 
& = \max_{\vx_n, q(\xi_n)}  \tfrac{2}{N}|\langle \xi_n  \rangle_{q(\xi_n)} \mbox{vec}(\vx_n \vx_n\trp)|_2 \leq \tfrac{1}{2N},  
\end{align} due to \eqref{gen_sen} and the fact that  
the mean of a PG variable $\langle \xi_n \rangle$ is given by 
\begin{align}
\int_{0}^\infty \xi_n \; \mbox{PG}(\xi_n|\; 1, \;\sqrt{\vx_n\trp \langle  \vm \vm\trp \rangle\vx_n}) d \xi_n 
= \frac{1}{2 \sqrt{\vx_n\trp \langle  \vm\vm\trp \rangle\vx_n}} \; \mbox{tanh} \left(\frac{\sqrt{\vx_n\trp \langle  \vm \vm\trp \rangle\vx_n}}{2} \right). \nonumber
\end{align} 
As shown in \figref{lambda}, the maximum value is $0.25$.

Since there are two perturbations (one for $\bar{\vs}_1$ and the other for  $\bar{\vs}_2$) in each iteration, 
we plug in $2J$ instead of the maximum iteration number $J$, when calculating the total privacy loss using the moments accountant.
Our VIPS algorithm for Bayesian logistic regression is given in Algorithm \ref{algo:VIPS_BLR}.

\begin{figure}[t]
\centering
\centerline{\includegraphics[width=0.9\textwidth]{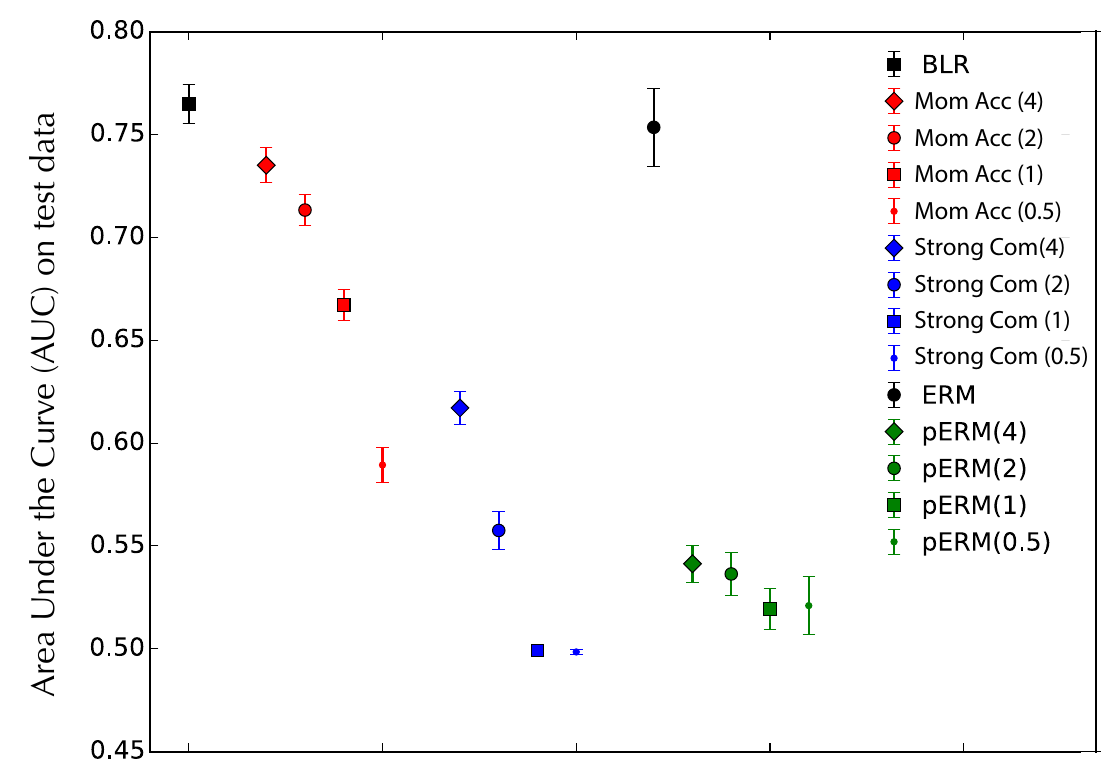}}
\caption{ Stroke data. Comparison between our method and private/non-private ERM, for different $\epsilon_{tot} \in \{0.5, 1, 2, 4 \}$. For the non-private methods, non-private BLR (black square marker) and ERM (black circle marker) achieved a similar AUC, which is higher than AUC obtained by any other private methods. 
Our method (red) under BLR with moments accountant with $\delta=0.0001$ achieved the highest AUCs regardless of $\epsilon_{tot}$ among all the other private methods. 
The private version of ERM (objective perturbation, green) performed worse than BRL with strong composition as well as BRL with moments accountant. While directly comparing these to the private ERM is not totally fair since the private ERM (pERM) is $\epsilon$-DP while others are ($\epsilon,\delta$)-DP, we show the difference between them in order to contrast the relative gain of our method compared to the existing method.  
}
\label{fig:Stroke_1}
\end{figure}

\subsection{Experiments with Stroke data}

We used the stroke dataset, which was first introduced by \citet{LethamRuMcMa14}
for predicting the occurrence of a stroke within a year after an atrial fibrillation diagnosis.\footnote{The authors extracted every patient in the MarketScan Medicaid Multi-State Database (MDCD) with a diagnosis of atrial fibrillation, one year of observation time prior to the diagnosis, and one year of observation time following the diagnosis.} 
There are $N=12,586$ patients in this dataset, and among these patients, 1,786 (14$\%$) had a stroke within a year of the atrial fibrillation diagnosis.


Following \cite{LethamRuMcMa14}, we also considered all drugs and all medical conditions of these patients as candidate predictors. A binary predictor variable is used for indicating the presence or absence of each drug or condition in the longitudinal record prior to the atrial fibrillation diagnosis. In addition, a pair of binary variables is used for indicating age and gender. These totalled $d=4,146$ unique features for medications and conditions. We randomly shuffled the data to make $5$ pairs of training and test sets. For each set, we used $10,069$ patients' records as training data and the rest as test data.

Using this dataset, we ran our VIPS algorithm in batch mode, i.e., using the entire training data in each iteration, as opposed to using a small subset of data. We also ran the private and non-private Empirical Risk Minimisation (ERM) algorithms \citep{ERM}, in which we performed $5$-fold cross-validation to set the regularisation constant given each training/test pair. As a performance measure, we calculated the \emph{Area Under the Curve} (AUC) on each test data, given the posteriors over the latent and parameters in case of BLR and the  parameter estimate in case of ERM.  In \figref{Stroke_1}, we show the mean and $1$-standard deviation of the AUCs obtained by each method.





\section{VIPS for sigmoid belief networks}\label{sigmoid_belief_nets}

\begin{SCfigure}[20][t]
\centering
\includegraphics[width=0.45\textwidth]{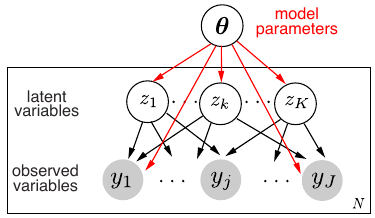}
  \caption{Schematic of the sigmoid belief network. A vector of binary observation $\vy =[y_1, \cdots, y_J]$ is a function of a vector of binary  latent variables $\vz = [z_1, \cdots, z_K]$ and model parameters such that $p(y_j=1|\vz, \vtheta) = \sigma(\vw_j\trp\vz+c_j)$. The latent variables are also binary such that $p(z_k=1|\vtheta) = \sigma(b_k)$,  where $\vtheta = \{\vw, \vc, \vb\}$.}
\label{fig:SBN}
\end{SCfigure}

As a last example model, we consider the Sigmoid Belief Network (SBN) model, which as first introduced in \cite{Neal92}, for modeling $N$ binary observations in terms of binary hidden variables and some parameters shown as \figref{SBN}. 
The complete-data likelihood for the $n$th observation and latent variables is given by
\begin{align}\label{eq:exact_data_like_SBN}
p(\Dat_n, \vl_n |\vm) &= p(\Dat_n|\vl_n, \vm) p(\vl_n|\vm), \nonumber \\
&= \prod_{j=1}^J \sigma(\vw_j\trp\vz_n + c_j) \prod_{k=1}^K \sigma(b_k), \\
&=\prod_{j=1}^J \frac{[\exp(\vw_j\trp\vz_n + c_j)]^{y_{n,j}}}{1+\exp(\vw_j\trp\vz_n + c_j)}
\prod_{k=1}^K \frac{[\exp(b_k)]^{z_{n,k}}}{1+\exp(b_k)},
\end{align} where we view the latent variables $\vl_n=\vz_n$
and model parameters $\vm=\vtheta$, and each datapoint $\Dat_n=\vy_n$.  

%
%

%

%
Thanks to the P{\'o}lya-Gamma data augmentation strategy, we can rewrite the complete-data likelihood from
 \begin{eqnarray}
p(\vz_n, \vy_n |\vtheta)
= \prod_{j=1}^J \frac{[\exp(\psi_{n,j})]^{y_{n,j}}}{1+\exp(\psi_{n,j})}
\prod_{k=1}^K \frac{[\exp(b_k)]^{z_{n,k}}}{1+\exp(b_k)}, 
\end{eqnarray} where we denote $\psi_{n,j} = \vw_j\trp\vz_n + c_j$, 
to 
 \begin{eqnarray}
p(\vy_n, \vz_n |\vxi^{(0)}_n, \vxi^{(1)}, \vtheta) \propto 
 \prod_{j=1}^J\exp((y_{n,j}-\tfrac{1}{2})\psi_{n,j})\exp(-\tfrac{\xi^{(0)}_{n,j} \psi_{n,j}^2}{2})
 \prod_{k=1}^K
 \exp((z_{n,k}-\tfrac{1}{2})b_k)\exp(-\tfrac{\xi^{(1)}_{k} b_k^2}{2}) \nonumber 
\end{eqnarray} where each element of vectors $\vxi^{(0)}_i \in \mathbb{R}^{J}$ and $\vxi^{(1)}\in \mathbb{R}^{K}$ is from PG($1,0$).

Using the notations $\vpsi_n \in \mathbb{R}^J$ where $\vpsi_n = W\vz_n + \vc$, $W = [\vw_1,  \cdots , \vw_J]\trp \in \mathbb{R}^{J \times K}$, and $\vone_J$ is a vector of $J$ ones, we obtain
 \begin{align}
p(\vy_n, \vz_n |\vxi^{(0)}_n, \vxi^{(1)}, \vtheta) \propto 
\exp \left[(\vy_{n}-\tfrac{1}{2}\vone_J)\trp \vpsi_{n}-\tfrac{1}{2}\vpsi_{n} \trp \mbox{diag}(\vxi^{(0)}_{n}) \vpsi_{n}
+(\vz_{n}-\tfrac{1}{2}\vone_K)\trp \vb -\tfrac{1}{2}\vb\trp \mbox{diag}(\vxi^{(1)}) \vb \right]. \nonumber 
\end{align} The complete-data likelihood given the PG variables provides the exponential family form  
\begin{align}
& p(\vy_n, \vz_n |\vxi^{(0)}_n, \vxi^{(1)}, \vtheta) \\
& \quad \propto \exp[(\vy_{n}-\tfrac{1}{2}\vone_J)\trp(W\vz_n + \vc) -\tfrac{1}{2}(W\vz_n+ \vc)\trp \mbox{diag}(\vxi^{(0)}_{n})(W\vz_n + \vc) \nonumber \\
& \qquad \qquad + (\vz_{n}-\tfrac{1}{2}\vone_K)\trp \vb -\tfrac{1}{2}\vb\trp \mbox{diag}(\vxi^{(1)}) \vb  ], \\
& \quad \propto  \exp[\vn(\vtheta)\trp \vs(\vy_n, \vz_n, \vxi^{(0)}_n, \vxi^{(1)}) ],
\end{align} where the natural parameters and sufficient statistics are given by 
\begin{equation}
 \vn(\vtheta) = \begin{bmatrix}
       \vb     \\[0.3em]
       - \frac{1}{2}\mbox{vec}(\vb \vb\trp)            \\[0.3em]
              \vc     \\[0.3em]
       - \frac{1}{2}\mbox{vec}(\vc \vc\trp)            \\[0.3em]
       - \frac{1}{2} \mbox{vec}( \mbox{diag}(\vc) W)     \\[0.3em]
       \mbox{vec}(W)           \\[0.3em]
       -\frac{1}{2} \mbox{vec}(\mbox{vec}(W\trp) \mbox{vec}(W\trp)\trp) \\[0.3em]
     \end{bmatrix}, 
     \quad 
     \vs(\vy_n, \vz_n, \vxi_n^{(0)}, \vxi^{(1)}) =  \begin{bmatrix}
        \vz_n - \frac{1}{2} \vone_K  \\[0.3em]
       \mbox{vec}(\mbox{diag}(\vxi^{(1)}))       \\[0.3em]
       \vy_n - \frac{1}{2} \vone_J   \\[0.3em]
       \mbox{vec}(\mbox{diag}(\vxi_n^{(0)}))       \\[0.3em]
       \mbox{vec}(\vxi_n^{(0)} \vz_n \trp)        \\[0.3em]
       \mbox{vec}(\vz_n (\vy_n - \frac{1}{2} \vone_J)\trp) \\[0.3em]
       \mbox{vec}(\mbox{diag}(\vxi_n^{(0)}) \otimes (\vz_n\vz_n\trp))  \\[0.3em]
            \end{bmatrix}.\nonumber
\end{equation}

\begin{algorithm}[t]
\caption{VIPS for sigmoid belief networks}
\label{algo:VIPS_SBN}
\begin{algorithmic}
\REQUIRE Data $\Dat$. Define $\rho_{t} = (\tau_0 + t)^{-\kappa}$, mini-batch size $S$, maximum iterations $T$, and $\sigma$
\ENSURE Privatised expected natural parameters $\tilde{\bar{\vn}}$ and expected sufficient statistics $\tilde{\bar{\vs}}$
\FOR{$t = 1, \ldots, T $}
\STATE {\it{\textbf{(1) E-step}}}: Given expected natural parameters $\bar{\vn}$, compute $q(\vxi^{(0)}_n)$
for $n=1,\ldots,S$. \\
Given $\bar{\vn}$, compute $q(\vxi^{(1)})$
for $n=1,\ldots,S$. \\
Given $\bar{\vn}$, $q(\vxi^{(0)}_n)$ and $q(\vxi^{(1)})$, compute $q(\vz_n)$
for $n=1,\ldots,S$. \\
Perturb $\bar{\vs} = \tfrac{1}{S} \sum_{n=1}^S \langle \vs( \Dat_n, \vxi^{(0)}_n, \vxi^{(1)}) \rangle_{q(\vxi^{(0)}_n) q(\vxi^{(1)}) q(\vz)}$ by Appendix B, and output $ \tilde{\bar{\vs}} $.
\STATE Update the moments function.
\STATE {\it{\textbf{(2) M-step}}}: Given $ \tilde{\bar{\vs}} $,
compute $q(\vm)$ by $\tilde{\vnu}^{(t)} = \vnu + N \tilde{\bar{\vs}} $.
Set $\tilde{\vnu}^{(t)} \mapsfrom (1-\rho_t)\tilde{\vnu}^{(t-1)} + \rho_t \tilde{\vnu}^{(t)} $.
Using $\tilde{\vnu}^{(t)}$, update variational posteriors for hyper-priors by Appendix C, and output $\tilde{\bar{\vn}} = \langle \vm \rangle_{q(\vm)}$.
\ENDFOR
\end{algorithmic}
\end{algorithm}

Now the PG variables form a set of sufficient statistics, which is separated from the model parameters. 
Similar to logistic regression, in the E-step, we compute the posterior over $\vxi$ and $\vz$, and output perturbed expected sufficient statistics.
The closed-form update of the posteriors over the PG variables is simply
\begin{eqnarray}
q(\vxi_{n}^{(0)}) &=& \prod_{j=1}^J q(\vxi_{n, j}^{(0)}) =\prod_{j=1}^J  \mbox{PG}(1, \sqrt{\langle(\vw_j\trp \vz_n + c_j)^2 \rangle_{q(\vtheta) q(\vz_n)}}), \\
q(\vxi^{(1)})&=& \prod_{k=1}^K q(\vxi_k^{(1)}) = \prod_{k=1}^K \mbox{PG}(1, \sqrt{\langle b_k^2 \rangle_{q(\vtheta)}}).
\end{eqnarray} The posterior over the latent variables is given by 
\begin{eqnarray}\label{eq:post_z}
q(\vz) &=& \prod_{n=1}^N \prod_{k=1}^K q(z_{n,k}) = \mbox{Bern}(\sigma(d_{n,k})), \\
d_{n,k}&=& \langle b_k \rangle_{q(\vtheta)} + \langle \vw_k\trp\vy_n\rangle_{ q(\vtheta)} -\tfrac{1}{2} \sum_{j=1}^J ( \langle w_{j,k} \rangle_{q(\vtheta)}
+ \langle \xi_{n,j}^{(0)}\rangle_{q(\vxi)}  [2 \langle \psi_{n,j}^{\setminus k} w_{j,k}\rangle_{q(\vtheta) q(\vz)}  +\langle w_{j,k}^2 \rangle_{q(\vtheta)} ]                     ), \nonumber \\
 \psi_{n,j}^{\setminus k} &=& \vw_j\trp\vz_n - w_{j,k} z_{n,k} + c_j.
\end{eqnarray}
Now, using $q(\vz)$ and $q(\vxi)$, we compute the expected sufficient statistics,
\begin{equation}
     \bar{\vs}(\Dat)
     =  \begin{bmatrix}
       \bar{\vs}_1 = \frac{1}{N}\sum_{n=1}^N  \vs_1(\vy_n)  \\[0.3em]
       \bar{\vs}_2 =  \frac{1}{N}\sum_{n=1}^N \vs_2(\vy_n)  \\[0.3em]
        \bar{\vs}_3 = \frac{1}{N}\sum_{n=1}^N \vs_3(\vy_n)  \\[0.3em]
       \bar{\vs}_4 =  \frac{1}{N}\sum_{n=1}^N \vs_4(\vy_n)  \\[0.3em]
        \bar{\vs}_5 = \frac{1}{N}\sum_{n=1}^N \vs_5(\vy_n)  \\[0.3em]
       \bar{\vs}_6 =  \frac{1}{N}\sum_{n=1}^N \vs_6(\vy_n)  \\[0.3em]
        \bar{\vs}_7 = \frac{1}{N}\sum_{n=1}^N \vs_7(\vy_n)  \\[0.3em]
            \end{bmatrix}, 
            \quad \mbox{ where }
 \begin{bmatrix}
       {\vs}_1(\vy_n)  \\[0.3em]
       {\vs}_2(\vy_n)  \\[0.3em]
       {\vs}_3(\vy_n)  \\[0.3em]
       {\vs}_4(\vy_n)  \\[0.3em]
       {\vs}_5(\vy_n)  \\[0.3em]
       {\vs}_6(\vy_n)  \\[0.3em]
       {\vs}_7(\vy_n)  \\[0.3em]
            \end{bmatrix} 
            =
     \begin{bmatrix}
        \langle \vz_n \rangle - \frac{1}{2} \vone_K  \\[0.3em]
        \mbox{vec}(\mbox{diag}(\langle \vxi^{(1)} \rangle))       \\[0.3em]
      \vy_n - \frac{1}{2} \vone_J  \\[0.3em]
      \mbox{vec}(\mbox{diag}(\langle \vxi_n^{(0)} \rangle ) )    \\[0.3em]
      \mbox{vec}(  \langle \vxi_n^{(0)} \rangle \langle \vz_n \rangle\trp) \\[0.3em]
    \mbox{vec}(\langle \vz_n \rangle (\vy_n - \frac{1}{2} \vone_J)\trp) \\[0.3em]
    \mbox{vec}( \mbox{diag}(\langle \vxi_n^{(0)} \rangle) \otimes (\langle \vz_n \vz_n\trp \rangle))  \\[0.3em]
            \end{bmatrix}.\nonumber
\end{equation}
Using the variational posterior distributions in the E-step, we perturb and output these sufficient statistics. See Appendix B for the sensitivities of each of these sufficient statistics. Note that when using the subsampled data per iteration,  the sensitivity analysis has to be modified as now the query is evaluated on a smaller dataset. Hence, the $1/N$ factor has to be changed to $1/S$.


Note that the fact that each of these sufficient statistics has a different sensitivity makes it difficult to directly use the composibility theorem of the moments accountant method \citep{2016arXiv160700133A}. To resolve this, we modify the sufficient statistic vector into a new vector with a fixed sensitivity, and then apply the Gaussian mechanism. In this case, the log-moments are additive since the Gaussian noise added to the modified vector of sufficient statistics given each subsampled data is independent, and hence we are able to perform the usual moments accountant composition.  Finally, we recover an estimate of the sufficient statistics from the perturbed modified vector.

In more detail, let us denote the sensitivities of each vector quantities by $C_1, \cdots, C_7$. Further, denote the moments accountant noise parameter by $\sigma$, and the subsampled data by $\Dat_q$ with sampling rate $q$.
We first scale down each sufficient statistic vector by its own sensitivity, so that the concatenated vector (denoted by $\vs'$ below)'s sensitivity becomes $\sqrt{7}$. Then, add the standard normal noise to the vectors with scaled standard deviation, $\sqrt{7}\sigma$. We then scale up each perturbed quantities by its own sensitivity,
given as 
\begin{align}
     \begin{bmatrix}
       \tilde{\vs}_1(\Dat_q)  \\[0.3em]
       \tilde{\vs}_2(\Dat_q)  \\[0.3em]
       \tilde{\vs}_3(\Dat_q)  \\[0.3em]
       \tilde{\vs}_4(\Dat_q)  \\[0.3em]
       \tilde{\vs}_5(\Dat_q)  \\[0.3em]
       \tilde{\vs}_6(\Dat_q)  \\[0.3em]
       \tilde{\vs}_7(\Dat_q)  \\[0.3em]
    \end{bmatrix} 
    =
     \begin{bmatrix}
       C_1 \tilde\vs'_1  \\[0.3em]
       C_2 \tilde\vs'_2 \\[0.3em]
       C_3 \tilde\vs'_3  \\[0.3em]
       C_4 \tilde\vs'_4 \\[0.3em]
       C_5 \tilde\vs'_5  \\[0.3em]
       C_6 \tilde\vs'_6 \\[0.3em]
        C_7 \tilde\vs'_7 \\[0.3em]
    \end{bmatrix} 
    \mbox{ where }
    \tilde{\vs}' = \vs'    + \sqrt{7}\sigma \Nrm(0,I), 
\end{align} where the $i$th chunk of the vector $\vs$ is $\vs'_i = {\bar{\vs}}_i/C_i$, resulting in privatized sufficient statistics.



\begin{figure}[t]
\centering
\centerline{\includegraphics[width=0.6\textwidth]{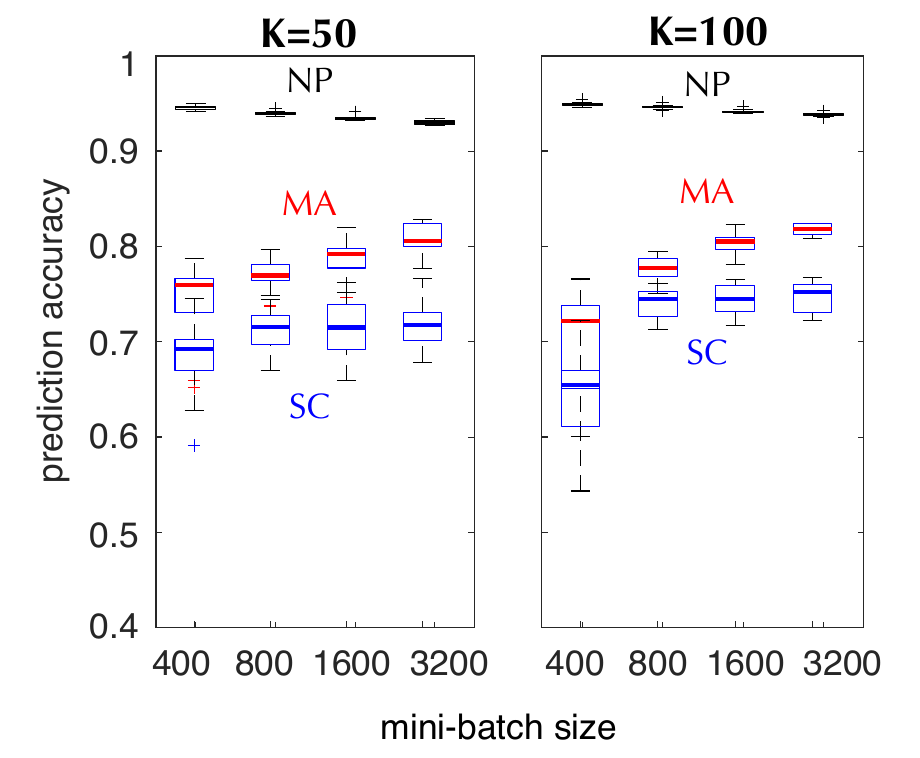}}
\caption{ Prediction accuracy (binarised version of MNIST dataset). 
%
%
The non-private version (NP in black) achieves highest prediction accuracy under the SBN model with $K=100$. 
The private versions (moments accountant : MA (in red), and strong composition : SC (in blue)) with $\sigma = 1$ and $K=100$ resulted in a total privacy loss $\epsilon_{tot}=2.3468$ when the mini-batch size is $S=400$, and $\epsilon_{tot}=2.398$ for $S=800$, $\epsilon_{tot}=3.2262$ for $S=1600$, and $\epsilon_{tot}=4.8253$ for $S=3200$. We ran these algorithms until they see the total training data, resulting in different numbers of iterations for each minibatch size. 
The private version using strong composition (blue) performs worse than that using moments accountant (red) regardless of the size of mini-batches, since the level of additive noise per iteration in strong composition is higher than that in  moments accountant.  
}
\label{fig:SBN_K}
\end{figure}

While any conjugate priors are acceptable, following \cite{GanHCC15}, we put a Three Parameter Beta Normal (TPBN) prior for $W$ 
\begin{eqnarray}
W_{j,k} \sim \Nrm(0, \zeta_{j,k}), \zeta_{j,k} \sim \mbox{Gam}(\tfrac{1}{2}, \xi_{j,k} ), \xi_{j,k} \sim \mbox{Gam}(\tfrac{1}{2}, \phi_k), \phi_k \sim \mbox{Gam}(0.5, \omega), \omega \sim \mbox{Gam}(0.5, 1) \nonumber
\end{eqnarray}to induce sparsity,  and isotropic normal priors for $\vb$ and $\vc$: $\vb \sim \Nrm(0, \nu_b I_K), \vc \sim \Nrm(0, \nu_c I_J)$, assuming these hyperparameters are set such that the prior is broad.
The M-step updates for the variational posteriors for the parameters as well as for the hyper-parameters are given in Appendix C. It is worth noting that these posterior updates for the hyper-parameters are one step away from the data, meaning that the posterior updates for the hyper-parameters are functions of variational posteriors that are already perturbed due to the perturbations in the expected natural parameters and expected sufficient statistics. Hence, we do not need any additional perturbation in the posteriors for the hyper-parameters.


\clearpage
\subsection{Experiments with MNIST data}
We tested our VIPS algorithm for the SBN model on the MNIST digit dataset which contains 60,000 training images of ten handwritten digits (0 to 9), where each image consists of $28 \times 28$ pixels. For our experiment, we considered a one-hidden layer SBN with $100$ hidden units $K= 100$.\footnote{We chose this number since when $K$ is larger than $100$, the variational lower bound on the test data, shown in Figure 2 in  \citep{GanHCC15}, does not increase significantly.} We varied the mini-batch size $S=\{400, 800, 1600, 3200\}$. For a fixed $\sigma=1$, we obtained two different values of privacy loss due to different mini-batch sizes. We ran our code until it sees the entire training data ($60,000$). 
%
We tested the non-private version of VIPS as well as the private versions with strong and moments accountant compositions, where each algorithm was tested in $10$ independent runs with different seed numbers. 

As a performance measure, we calculated the pixel-wise prediction accuracy by first converting the pixels of reconstructed images into probabilities (between 0 and 1); then converting the probabilities into binary variables; and averaging the squared distances between the predicted binary variables and the test images. 
In each seed number, we selected $100$ randomly selected test images from $10,000$ test datapoints. \figref{SBN_K} shows the performance of each method in terms of prediction accuracy. 


\section{Discussion}\label{Discussion}

We have developed a practical privacy-preserving VB algorithm which outputs accurate and privatized expected sufficient statistics and expected natural parameters.  Our approach uses the moments accountant analysis combined with the privacy amplification effect due to subsampling of data, which significantly decrease the amount of additive noise for the same expected privacy guarantee compared to the standard analysis. 
Our methods show how to perform  variational Bayes inference in private settings, not only for the conjugate exponential family models but also for non-conjugate models with binomial likelihoods using the Poly{\'a} Gamma data augmentation. We illustrated the effectiveness of our algorithm on several real-world datasets. 

The private VB algorithms for the Latent Dirichlet Allocation (LDA), Bayesian Logistic Regression (BLR), and Sigmoid Belief Network (SBN) models we discussed are just a few examples of a much broader class of models to which our private VB framework applies. Our positive empirical results with VB indicate that these ideas are also likely to be beneficial for privatizing many other iterative machine learning algorithms. 
In future work, we plan to apply this general framework to other inference methods for larger and more complicated models such as deep neural networks. More broadly, our vision is that \emph{practical} privacy preserving machine learning algorithms will have a transformative impact on the practice of data science in many real-world applications.

\newpage
\bibliography{DPEPrefs}

\newpage

\section*{Appendix A : Variational lower bound with auxiliary variables}

The variational lower bound (per-datapoint for simplicity) given by 
\begin{align}\label{eq:aux_lbd}
\mathcal{L}_n (q(\vm), q(\vl_n)) &= \int \; q(\vm) q(\vl_n) \log \frac{p(\Dat_n | \vl_n, \vm)p(\vm) p(\vl_n)}{q(\vm) q(\vl_n)} d\vl_n  d\vm, \\
&= \int \; q(\vm) q(\vl_n) \log p(\Dat_n | \vl_n, \vm) d\vl_n  d\vm - \mbox{D}_{KL}(q(\vm)||p(\vm)) - \mbox{D}_{KL}(q(\vl_n)||p(\vl_n)), \nonumber 
\end{align} 
where the first term can be re-written using \eqref{PG_AUG}, 
\begin{align}\label{eq:aux_lbd_likelihood}
\int \; q(\vm) q(\vl_n) \log p(\Dat_n | \vl_n, \vm) d\vl_n  d\vm
&= -b \log 2 + (y_n - \tfrac{b}{2}) \langle \vl_n \rangle_{q(\vl_n)} \trp \langle \vm \rangle_{q(\vm)} \nonumber \\
&  \quad + \int \; q(\vm) q(\vl_n) \log \int_0^\infty \exp(-\tfrac{1}{2}\xi_n\vl_n\trp \vm \vm\trp \vl_n) p(\xi_n) d\xi_n. \nonumber
\end{align} We rewrite the third term using $q(\xi_n)$, the variational posterior distribution for $\xi_n$,
\begin{align}
& \int \; q(\vm) q(\vl_n) \log \int_0^\infty \exp(-\tfrac{1}{2}\xi_n\vl_n\trp \vm \vm\trp \vl_n) p(\xi_n) d\xi_n \nonumber \\
&= \int \; q(\vm) q(\vl_n) \log \int_0^\infty q(\xi_n) \exp(-\tfrac{1}{2}\xi_n\vl_n\trp \vm \vm\trp \vl_n) \frac{p(\xi_n)}{q(\xi_n)} d\xi_n, \nonumber \\
&\geq  \int \; q(\vm) q(\vl_n) \int_0^\infty q(\xi_n) \left[ (-\tfrac{1}{2}\xi_n\vl_n\trp \vm \vm\trp \vl_n) +  \log  \frac{p(\xi_n)}{q(\xi_n)} \right] d\xi_n, \nonumber \\
& \quad = -\tfrac{1}{2} \langle \xi_n \rangle_{q(\xi_n)} \langle \vl_n\trp \vm \vm\trp \vl_n \rangle_{q(\vl_n)q(\vm)} -  \mbox{D}_{KL}(q(\xi_n)||p(\xi_n)),
\end{align} which gives us a lower lower bound to the log likelihood, 
\begin{align}
\mathcal{L}_n (q(\vm), q(\vl_n)) 
&\geq \mathcal{L}_n (q(\vm), q(\vl_n), q(\xi_n)) \\
& \quad := -b \log 2 + (y_n - \tfrac{b}{2}) \langle \vl_n \rangle_{q(\vl_n)} \trp \langle \vm \rangle_{q(\vm)} -\tfrac{1}{2} \langle \xi_n \rangle_{q(\xi_n)} \langle \vl_n\trp \vm \vm\trp \vl_n \rangle_{q(\vl_n)q(\vm)} , \nonumber \\
& \qquad -  \mbox{D}_{KL}(q(\xi_n)||p(\xi_n)) - \mbox{D}_{KL}(q(\vm)||p(\vm)) - \mbox{D}_{KL}(q(\vl_n)||p(\vl_n)), \nonumber 
\end{align} which implies that
\begin{align}
&\int \; q(\vm) q(\vl_n) q(\xi_n) \log p(\Dat_n|\vl_n, \xi_n, \vm) \nonumber \\
& = 
 -b \log 2 + (y_n - \tfrac{b}{2}) \langle \vl_n \rangle_{q(\vl_n)} \trp \langle \vm \rangle_{q(\vm)} -\tfrac{1}{2} \langle \xi_n \rangle_{q(\xi_n)} \langle \vl_n\trp \vm \vm\trp \vl_n \rangle_{q(\vl_n)q(\vm)}.
\end{align}

\newpage

\section*{Appendix B : Perturbing expected sufficient statistics in SBNs}

Using the variational posterior distributions in the E-step, we perturb and output each sufficient statistic as follows. Note that the $1/N$ factor has to be changed to $1/S$ when using the subsampled data per iteration. 
\begin{itemize}
\item 
For perturbing ${\bar{\vs}}_1 $, we perturb $A = \frac{1}{N} \sum_{n=1}^N\langle \vz_n \rangle$
where the sensitivity is given by 
\begin{align}
\Delta A &= \max_{|\Dat \setminus {\Dat}'|=1} |A(\Dat)-A({\Dat}')|_2, \nonumber \\
&\leq  \max_{\vy_n, q(\vz_n)} \frac{1}{N} \sqrt{ \sum_{k=1}^K (\sigma(d_{n,k}) )^2}
 \leq  \frac{\sqrt{K}}{N},
\end{align} due to \eqref{gen_sen} and the fact that $\vz_n$ is a vector of Bernoulli random variables (length $K$) and the mean of each element is  $\sigma(d_{n,k})$ as given in \eqref{post_z}.
\item For perturbing $\tilde{\bar{\vs}}_2$, we perturb $B = \langle \vxi^{(1)} \rangle $, i.e. the part before we apply the diag and vec operations. 
The sensitivity of $B$ is given by  $\Delta B \leq \frac{\sqrt{K}}{4}$ as shown in \figref{lambda}.
%
\item For perturbing $\bar{\vs}_3$, we perturb $C = \frac{1}{N}\sum_{n=1}^N \vy_n $, where the sensitivity is given by 
%
\begin{eqnarray}
\Delta C= \max_{|\Dat \setminus \Dat'|=1} |C(\Dat)-C(\Dat')|_2 = \max_{\vy_n} \tfrac{1}{N} | \vy_n|_2 \leq  \tfrac{\sqrt{J}}{N},
\end{eqnarray} since  $\vy_n$ is a binary vector of length $J$. Note that when performing the batch optimisation, we perturb $\bar{\vs}_3$ only once, since this quantity remains the same across iterations. However, when performing the stochastic optimisation, we perturb $\bar{\vs}_3$ in every iteration, since the new mini-batch of data is selected in every iteration.
\item For perturbing $\tilde{\bar{\vs}}_4$, we perturb $D=\frac{1}{N}\sum_{n=1}^N\langle \vxi_n^{(0)}\rangle $, which is once again the part before taking diag and vec operations. 
Due to \eqref{gen_sen}, the sensitivity is given by
\begin{align}
\Delta D &= \max_{|\Dat \setminus \Dat'|=1} |D(\Dat)-D(\Dat')|_2 \leq  \frac{\sqrt{J}}{4N}.
\end{align} 
%
\item For $\tilde{\bar{\vs}}_5$, we perturb $E = \frac{1}{N}\sum_{n=1}^N \langle \vxi_n^{(0)} \rangle  \langle \vz_n \rangle \trp$
.  From \eqref{gen_sen}, the sensitivity is given by
\begin{align}
\Delta E &= \max_{|\Dat  \setminus \Dat'|=1} |E(\Dat)-E(\Dat')|_2, \nonumber \\
& \leq \max_{\vy_n, q(\vz_n), q(\xi_n)} \frac{1}{N}\sqrt{ \sum_{k=1}^K \sum_{j=1}^J (\langle \vxi_{n, j}^{(0)} \rangle  \langle \vz_{n,k} \rangle  )^2} \leq  \frac{\sqrt{JK}}{4N}.
\end{align} 
\item For $\tilde{\bar{\vs}}_6$, we can use the noisy  $\tilde{\bar{\vs}}_1$ for the second term, but perturb only the first term $F = \frac{1}{N}\sum_{n=1}^N \vy_n \langle\vz_n \rangle\trp$. 
Due to \eqref{gen_sen}, the sensitivity is given by
\begin{align}
\Delta F &= \max_{|\Dat \setminus \Dat'|=1} |F(\Dat)-F(\Dat')|_2, \nonumber \\
& \leq \max_{\vy_n, q(\vz_n)} \frac{1}{N}\sqrt{\sum_{k=1}^K \sum_{j=1}^J (\vy_{n, j}  \langle \vz_{n,k} \rangle )^2} \leq  \frac{\sqrt{JK}}{N}.
\end{align} 
\item For ${\bar{\vs}}_7$, we define a matrix $G$, which is a collection of $J$ matrices where each matrix is $G_j = \frac{1}{N}\sum_{n=1}^N \langle \vxi_{n,j}^{(0)} \rangle \langle \vz_n \vz_n\trp\rangle $, where $G_j \in \mathbb{R}^{K \times K}$. 

Using \eqref{gen_sen}, the sensitivity of $G_j$ is given by
\begin{align}
\Delta G_j &= \max_{|\Dat \setminus \Dat'|=1} |G_j(\Dat)- G_j(\Dat')|_2, \nonumber \\
&\leq  \max_{\vy_n} \tfrac{1}{N} \sqrt{ \sum_{k=1}^K \sum_{k'=1}^K  (\langle \vxi_{n,j}^{(0)} \rangle \langle \vz_{n,k}   \vz_{n,k} \rangle )^2}
 \leq  \tfrac{K}{4N},
\end{align}
which gives us the sensitivity of $\Delta G \leq \frac{\sqrt{J}K}{4N}$.

\end{itemize}

\newpage 
\section*{Appendix C: M-step updates in SBNs}

The M-step updates are given below (taken from \citep{GanHCC15}):
\begin{itemize}
%
\item $q(W) = \prod_{j=1}^J q(\vw_j)$, and $q(\vw_j) = \Nrm(\mu_j, \Sigma_j)$, where 
\begin{align}
\Sigma_j &= \left[ \sum_{n=1}^N \langle \xi_{n,j}^{(0)} \rangle_{q(\vxi)} \langle \vz_n \vz_n\trp\rangle  + \mbox{diag}(\langle \vzeta_j^{-1}\rangle_{q(\vzeta)} ) \right]^{-1} \nonumber \\
 \vmu_j &= \Sigma_j \left[ \sum_{n=1}^N (\vy_{n,j} - \tfrac{1}{2} - \langle c_j \rangle_{q(\vtheta)} \langle \xi_{n,j}^{(0)} \rangle_{q(\vxi)}       ) \langle \vz_n\rangle_{q(\vz)}   \right]
 .\nonumber
\end{align} where we replace $\frac{1}{N}\sum_{n=1}^N \langle \xi_{n,j}^{(0)} \rangle_{q(\vxi)} \langle \vz_n \vz_n\trp\rangle $,  $\frac{1}{N}\sum_{n=1}^N \vy_{n,j}\langle \vz_n\rangle_{q(\vz)}$,  $\frac{1}{N}\sum_{n=1}^N\langle \vz_n\rangle_{q(\vz)}$, and $\frac{1}{N}\sum_{n=1}^N\langle \xi_{n,j}^{(0)} \rangle_{q(\vxi)} \langle \vz_n\rangle_{q(\vz)} $, with  perturbed expected sufficient statistics.
\item $q(\vb) = \Nrm(\vmu_\vb, \Sigma_\vb)$, where
\begin{eqnarray}
\Sigma_\vb &=& \left[ \frac{1}{\nu_b} I + N\mbox{diag}(\langle \vxi^{(1)} \rangle) \right]^{-1}, \\
\vmu_\vb &=& \Sigma_\vb \left[ \sum_{n=1}^N ( \langle \vz_n \rangle - \frac{1}{2} \vone_K ) \right],
\end{eqnarray} where we replace $\langle \vxi^{(1)} \rangle$ and $\frac{1}{N}\sum_{n=1}^N\langle \vz_n\rangle_{q(\vz)}$ with perturbed expected sufficient statistics.
\item $q(\vc) = \Nrm(\vmu_\vc, \Sigma_\vc)$, where
\begin{eqnarray}
\Sigma_\vc &=& \left[ \frac{1}{\nu_c} I + \mbox{diag}(\sum_{n=1}^N\langle \vxi_n^{(0)} \rangle) \right]^{-1}, \\
\vmu_\vc &=& \Sigma_\vc \left[ \sum_{n=1}^N ( \vy_n\trp - \frac{1}{2}\vone_J\trp - \frac{1}{2} \mbox{diag}(\langle \vxi_n^{(0)} \rangle\langle \vz_n \rangle\trp \langle W\rangle\trp) ) \right],
\end{eqnarray} where we replace $\frac{1}{N}\sum_{n=1}^N\langle \vxi_n^{(0)} \rangle$, $\frac{1}{N}\sum_{n=1}^N\vy_n$, and $\frac{1}{N} \sum_{n=1}^N \langle \vxi_n^{(0)} \rangle\langle \vz_n \rangle\trp$ with perturbed expected sufficient statistics.
\item TPBN shrinkage priors:
\begin{eqnarray}
q(\zeta_{j,k}) &=& \mathcal{GIG}(0, 2\langle \xi_{j,k} \rangle_{q(\xi)}, \langle w_{j,k}^2\rangle_{q(\vtheta)}), \; \; \qquad \quad q(\xi_{j,k}) \; = \; \mbox{Gam}(1, \langle \zeta_{j,k} \rangle_{q(\zeta)} ) \nonumber \\
q(\phi_k) &=&  \mbox{Gam}(\tfrac{J}{2}+ \tfrac{1}{2}, \langle \omega \rangle_{q(\omega)} + \sum_{j=1}^J \langle \xi_{j,k} \rangle_{q(\xi)}), \quad q(\omega) \; = \; \mbox{Gam}(\tfrac{K}{2}+ \tfrac{1}{2}, 1+ \sum_{k=1}^K \langle \phi_{k} \rangle_{q(\phi)}). \nonumber
\end{eqnarray}  When updating these TPBN shrinkage priors, we first calculate $q(\zeta_{j,k}) $ using a data-independent initial value for $2\langle \xi_{j,k} \rangle_{q(\xi)}$ and perturb $\langle w_{j,k}^2\rangle_{q(\vtheta)}$ (since $q(c)$ is perturbed). Using this perturbed $q(\zeta_{j,k})$, we calculate $q(\xi_{j,k}) $. Then, using $q(\xi_{j,k}) $, we calculate $q(\phi_k) $ with some data-independent initial value for $\langle \omega \rangle_{q(\omega)} $. Then, finally, we update  $q(\omega) $ using  $q(\phi_k) $. In this way, these TPBN shrinkage priors are perturbed.
\end{itemize}

\newpage 
\section*{Appendix D: The EM algorithm and its relationship to VBEM}
\label{app:EMvsVBEM}
In contrast to VBEM, which computes an approximate posterior distribution, EM finds a point estimate of model parameters $\vm$.
A derivation of EM from a variational perspective, due to \citet{neal1998view}, shows that VBEM generalizes EM \citep{Beal_03}.  
To derive EM from this perspective, we begin by identifying a lower bound $\mathcal{L}(q, \vm)$ on the log-likelihood using another Jensen's inequality argument, which will serve as a proxy objective function.  For any parameters $\vm$ and auxiliary distribution over the latent variables $q(\vl)$,
\begin{align}
  \log p(\Dat;\vm) &= 
   \log \Big (\int d\vl \; p(\vl,\Dat; \vm) \frac{q(\vl)}{q(\vl)} \Big )
   = \log \Big (\bbE_q \Big [\frac{p(\vl,\Dat; \vm)}{q(\vl)}\Big ] \Big) \nonumber \\
   &\geq \bbE_q \Big [\log p(\vl,\Dat;\vm) - \log q(\vl)\Big ] = \bbE_q \Big [\log p(\vl,\Dat;\vm)\Big ] + H(q) \triangleq \mathcal{L}(q, \vm) \mbox{.} \label{eq:EMbound}
\end{align}
Note that $\mathcal{L}(q, \vm)$ has a very similar definition to the ELBO from VB (\eqref{VB_ELBO_fromKL}).
Indeed, we observe that for an instance of VBEM where $q(\vm)$ is restricted to being a Dirac delta function at $\vm$, $q(\vm)= \delta(\vm - \vm^*)$, and with no prior on $\vm$, we have  $\mathcal{L}(q(\vl),\vm) = \mathcal{L}(q(\vl,\vm))$ \citep{Beal_03}.
The EM algorithm can be derived as a coordinate ascent algorithm which maximises $\mathcal{L}(q, \vm)$ by alternatingly optimizing $q$ (the E-step) and $\vm$ (the M-step) \citep{neal1998view}.  At iteration $i$, with current parameters $\vm^{(i-1)}$, the updates are:
\begin{align}
q^{(i)} &= \arg \max_{q}  \mathcal{L}(q, \vm^{(i-1)}) = p(\vl|\Dat; \vm^{(i-1)}) &\mbox{ (\textbf{E-step})\;} \nonumber \\
\vm^{(i)} &= \arg \max_{\vm} \mathcal{L}(q^{(i)}, \vm) &\mbox{ (\textbf{M-step}).} \label{eq:EM_variational}
\end{align}
In the above, $\arg \max_q  \mathcal{L}(q, \vm^{(i-1)}) = p(\vl|\Dat; \vm^{(i-1)})$ since this maximisation is equivalent to minimising $D_{KL}(q(\vl)\|p(\vl|\Dat; \vm^{(i-1)}))$.  To see this, we rewrite $\mathcal{L}(q,\vm)$ as
\begin{align}
\mathcal{L}(q,\vm) &= \bbE_q \Big [\log p(\vl,\Dat;\vm)\Big ] - \bbE_q[\log q(\vl)] \nonumber \\
&= \log p(\Dat;\vm) + \bbE_q \Big [\log p(\vl|\Dat, \vm) \Big ] - \bbE_q[\log q(\vl)] \nonumber \\
&= \log p(\Dat;\vm) - D_{KL}(q(\vl)\|p(\vl|\vm, \Dat))  \mbox{ .} \label{eq:EM_KL}
\end{align}
From a VBEM perspective, the M-step equivalently finds the Dirac delta $q(\vm)$ which maximises $\mathcal{L}(q(\vl,\vm))$.  Although we have derived EM as optimizing the variational lower bound $\mathcal{L}(q,\vm)$ in each E- and M-step, it can also be shown that it monotonically increases the log-likelihood $\log p(\Dat;\vm)$ in each iteration.  The E-step selects $q^{(i)} =p(\vl|\Dat; \vm^{(i-1)})$ which sets the KL-divergence in \eqref{EM_KL} to $0$  at $\vm^{(i)}$, at which point the bound is tight.  This means that $\log p(\Dat;\vm^{(i-1)})$ is achievable in the bound, so $\max_{\vm}\mathcal{L}(q^{(i)},\vm) \geq \log p(\Dat;\vm^{(i-1)})$.  As $\mathcal{L}(q,\vm)$ lower bounds the log likelihood at $\vm$, this in turn guarantees that the log-likelihood is non-decreasing over the previous iteration.  We can thus also interpret the E-step as computing a lower bound on the log-likelihood, and the M-step as maximizing this lower bound, i.e. an instance of the Minorise Maximise (MM) algorithm.  EM is sometimes written in terms of the \emph{expected complete data log-likelihood} $Q(\vm;\vm^{(i-1)})$,  
\begin{align}
Q(\vm;\vm^{(i-1)}) &= \bbE_{p(\vl|\Dat; \vm^{(i-1)})} [\log{p(\vl,\Dat;\vm)}] &\mbox{ (\textbf{E-step})\;}\nonumber \\
\vm^{(i)} &= \arg \max_{\vm} Q(\vm;\vm^{(i-1)}) &\mbox{ (\textbf{M-step}).} \label{eq:EM_traditional}
\end{align}
This is equivalent to \eqref{EM_variational}: $Q(\vm;\vm^{(i-1)}) + H(p(\vl|\Dat; \vm^{(i-1)})) = \mathcal{L}(p(\vl|\Dat; \vm^{(i-1)}),\vm)$.


\end{document}